\documentclass[sigconf]{acmart}

\AtBeginDocument{%
  \providecommand\BibTeX{{%
    \normalfont B\kern-0.5em{\scshape i\kern-0.25em b}\kern-0.8em\TeX}}}

\acmConference[p]{}
\acmBooktitle{}
\renewcommand\footnotetextcopyrightpermission[1]{}
\acmSubmissionID{2735}

\usepackage{bbding}
\usepackage{threeparttable}
\usepackage{graphicx}
\usepackage{subfigure}
\usepackage{booktabs}
\usepackage{amsmath}

\usepackage{stfloats}

\usepackage{multirow}

\usepackage{mathrsfs}
\begin{document}

%%
%% The "title" command has an optional parameter,
%% allowing the author to define a "short title" to be used in page headers.
\title{CrossHuman: Learning Cross-Guidance from Multi-Frame Images for Human Reconstruction}

%
% The "author" command and its associated commands are used to define
% the authors and their affiliations.
% Of note is the shared affiliation of the first two authors, and the
% "authornote" and "authornotemark" commands
% used to denote shared contribution to the research.
\author{Liliang Chen}\authornote{Both authors contributed equally to this research.}
% \email{trovato@corporation.com}
% \orcid{1234-5678-9012}
\affiliation{%
  \institution{OPPO Research Institute}
%   \streetaddress{P.O. Box 1212}
  \city{Beijing}
%   \state{Ohio}
  \country{China}
%   \postcode{43017-6221}
}
% \email{chenliliang1@oppo.com}

\author{Jiaqi Li}\authornotemark[1]
\affiliation{%
  \institution{Beihang University}
%   \streetaddress{P.O. Box 1212}
  \city{Beijing}
%   \state{Ohio}
  \country{China}
%   \postcode{43017-6221}
}
% \email{ljq2019@buaa.edu.cn}

\author{Han Huang}\authornote{Corresponding Author}
\affiliation{%
  \institution{OPPO Research Institute}
%   \streetaddress{P.O. Box 1212}
  \city{Beijing}
%   \state{Ohio}
  \country{China}
%   \postcode{43017-6221}
}
% \email{huanghan@oppo.com}

\author{Yandong Guo}

\affiliation{%
  \institution{OPPO Research Institute}
%   \streetaddress{P.O. Box 1212}
  \city{Beijing}
%   \state{Ohio}
  \country{China}
%   \postcode{43017-6221}
}
% \email{guoyandong@oppo.com}

%%
%% By default, the full list of authors will be used in the page
%% headers. Often, this list is too long, and will overlap
%% other information printed in the page headers. This command allows
%% the author to define a more concise list
%% of authors' names for this purpose.

% \renewcommand{\shortauthors}{Chen and Li, et al.}

%%
%% The abstract is a short summary of the work to be presented in the
%% article.
\begin{abstract}
  We propose CrossHuman, a novel method that learns cross-guidance from parametric human model and multi-frame RGB images to achieve high-quality 3D human reconstruction. To recover geometry details and texture even in invisible regions, we design a reconstruction pipeline combined with tracking-based methods and tracking-free methods. Given a monocular RGB sequence, we track the parametric human model in the whole sequence, the points (voxels) corresponding to the target frame are warped to reference frames by the parametric body motion. Guided by the geometry priors of the parametric body and spatially aligned features from RGB sequence, the robust implicit surface is fused. Moreover, a multi-frame transformer (MFT) and a self-supervised warp refinement module are integrated to the framework to relax the requirements of parametric body and help to deal with very loose cloth. Compared with previous works, our CrossHuman enables high-fidelity geometry details and texture in both visible and invisible regions and improves the accuracy of the human reconstruction even under estimated inaccurate parametric human models. The experiments demonstrate that our method achieves state-of-the-art (SOTA) performance.
\end{abstract}

%%
%% The code below is generated by the tool at http://dl.acm.org/ccs.cfm.
%% Please copy and paste the code instead of the example below.
%%
% \begin{CCSXML}
% <ccs2012>
% <concept>
% <concept_id>10010147.10010371.10010396</concept_id>
% <concept_desc>Computing methodologies~Shape modeling</concept_desc>
% <concept_significance>500</concept_significance>
% </concept>
% <concept>
% <concept_id>10010147.10010178.10010224.10010245.10010254</concept_id>
% <concept_desc>Computing methodologies~Reconstruction</concept_desc>
% <concept_significance>500</concept_significance>
% </concept>
% </ccs2012>
% \end{CCSXML}

% \ccsdesc[300]{Computing methodologies~Shape modeling}
% \ccsdesc[500]{Computing methodologies~Reconstruction}

%%
%% Keywords. The author(s) should pick words that accurately describe
%% the work being presented. Separate the keywords with commas.

%% A "teaser" image appears between the author and affiliation
%% information and the body of the document, and typically spans the
%% page.

%%
%% This command processes the author and affiliation and title
%% information and builds the first part of the formatted document.
\maketitle
\pagestyle{plain}
\begin{figure}[h]
\centering
\includegraphics[width=0.5\textwidth,trim={6cm 3cm 6cm 2cm},clip]{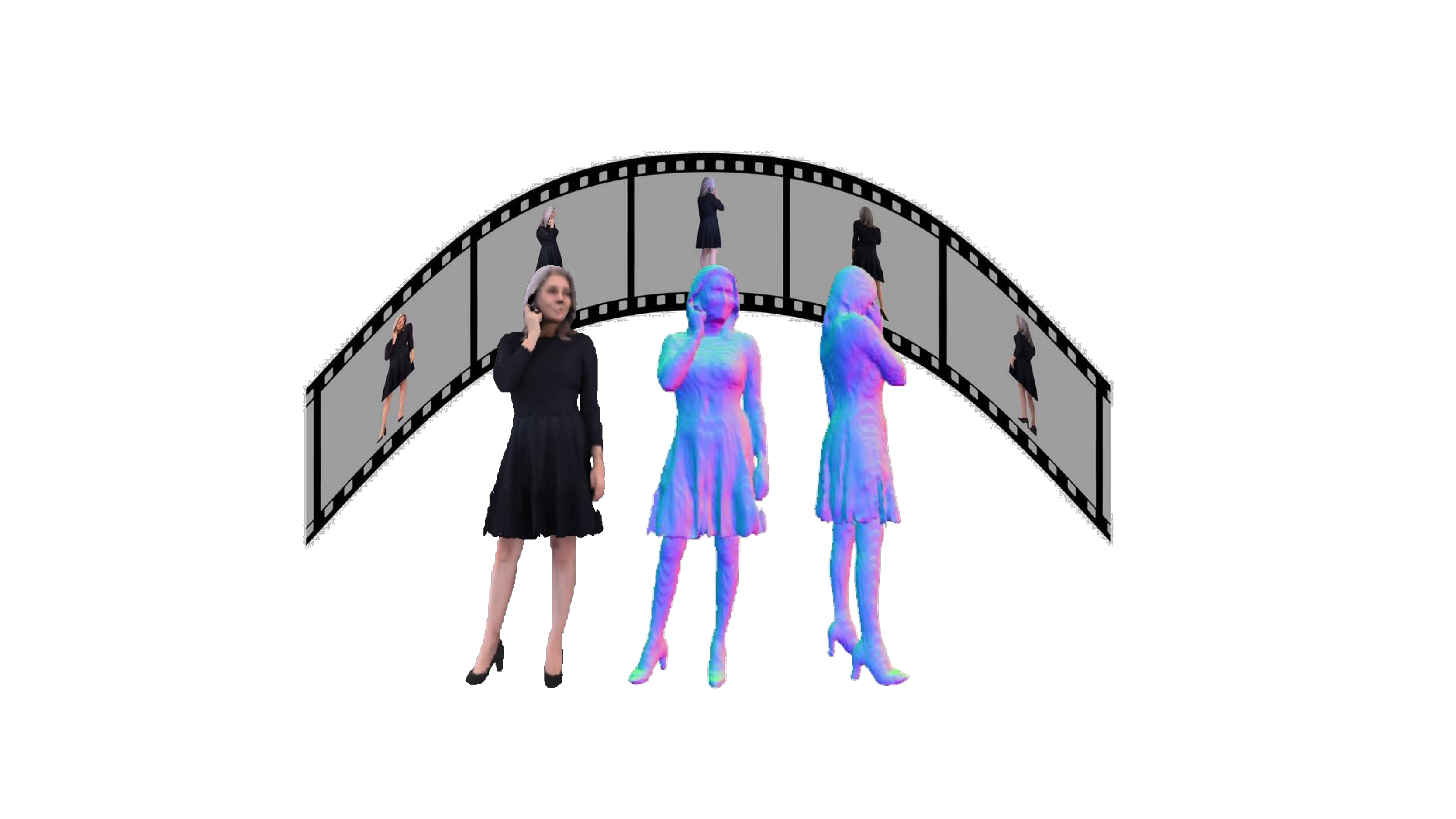}
\caption{The results reconstructed using our method. Given only a RGB sequence, our method is able to reconstruct complete human with rich details even under very loose cloth.}
\label{img1}
\end{figure}

\section{Introduction}
% no \IEEEPARstart
3D human reconstruction is a popular topic in computer vision and computer graphics, which is promising to enable various applications such as video games, holoportation and virtual try-on. To achieve surprisingly detailed geometry and texture reconstruction, existing methods \cite{leroy2017multi, gall2009motion, mustafa2015general, liu2013markerless, zhang2017detailed, newcombe2015dynamicfusion, yu2018doublefusion, xu2019flyfusion, alldieck2018video, guo2019relightables} often require multiple synchronized cameras, or a RGBD sensor. However, the expensive and professional setups limited their popularity. In contrast, single-view human reconstruction \cite{bogo2016keep, ma2020learning, kocabas2020vibe, zeng20203d, alldieck2019tex2shape, saito2019pifu, saito2020pifuhd, he2021arch++} have enabled the recovery of 3D human pose and shape estimation using a single camera, which becomes a hot research topic due to its low cost and convenient setup.

Many single-view human 3D reconstruction methods have been proposed. Among which, tracking-based methods \cite{habermann2019livecap, habermann2020deepcap, newcombe2015dynamicfusion, yu2018doublefusion, yu2017bodyfusion, zhi2020texmesh, alldieck2018detailed} utilize a statistical body template model as the geometrical prior and learn to deform the template, suffering from lack of fidelity and difficulty supporting clothing variations; while tracking-free methods \cite{saito2019pifu, saito2020pifuhd, chibane2020implicit, park2019deepsdf, chen2019learning, jiang2020local, genova2020local} using implicit surface functions can handle topological changes and reconstruct the 3D human body with high-fidelity details in visible regions. However, these methods can not deal with challenging poses and the surfaces in the invisible regions are usually over-smooth.

In the recent work POSEFusion \cite{li2021posefusion}, the authors propose a human volumetric capture method that combines tracking-based methods and tracking-free inference methods to achieve dynamic and high-fidelity 3D reconstruction from a RGBD camera. They use the skeleton tracking to obtain the pose and shape parameters of SMPL \cite{loper2015smpl} from depth stream, use pose-guided keyframe RGB image selection scheme to obtain more image details in visible and invisible regions, and finally fuse a complete human model. However, the RGBD camera is expensive and it is difficult for their method to deal with very loose cloth. In order to improve the accuracy of reconstruction system, the estimated parameters human model should be as accurate as possible. IP-Net \cite{bhatnagar2020combining} proposes a optimization based approach integrating viewpoint landmarks to optimize SMPL. PaMIR \cite{zheng2021pamir} designs a body reference optimization method to refine SMPL and uses the predicted SMPL at both the testing and the training stages. To a certain extent, such optimization methods can  fill the accuracy gap of SMPL between different stages, but the reconstruction quality is still strongly limited by the SMPL estimation.

State-of-the-art works of monocular SMPL estimation can output roughly correct pose and shape, but not very accurate. We believe that ordinary users' perception of the accuracy of reconstructing human body poses is relatively weak in many practical application scenarios. However, the slight SMPL pose estimation error will lead to many works that use SMPL priors to reconstruct the incomplete human body. The incomplete human reconstruction results are visually bad and will greatly affect the user experience. Therefore, we temporarily compromise on the insignificant pose error problem and try to find a way to both utilize the strong prior knowledge of SMPL model while avoiding the impact of SMPL error on reconstructed visual effects as much as possible. We hope to design a monocular reconstruction algorithm with practicality, integrity, realism, coherence and other elements of strong visual perception as primary goals.

To achieve high-quality geometry and texture reconstruction in both visible and invisible regions from a single RGB camera, in this paper we propose CrossHuman, which utilizes the estimated parametric human model and RGB images as guidance to reconstruct a complete human model. The geometry features of the SMPL serve as a coarse human shape proxy to guide our network handle challenging poses and encourage global shape regularity. Meanwhile, the pixel aligned features from multi-frame images lead to recover complex geometric and cloth wrinkles and resolve feature ambiguity. In addition, we further propose a new self-supervised training scheme in the CrossHuman-based framework, and achieve accurate reconstruction even under estimated inaccurate SMPL model. Our key contributions in this work can be summarized as follows: 

a)	A new single-view human 3d reconstruction pipeline that utilizes the parametric human model and RGB images as guidance, and achieves high-fidelity geometry and texture in both visible and invisible regions from a single RGB sequence.

b)	The first method that relaxes the requirement of the SMPL accuracy. Our method achieves high-quality human 3D reconstruction in cases of inaccurate SMPL estimations and challenging poses.

c)	A novel self-supervised training scheme that correct positions of warped points and a new transformer structure for multi-frame feature fusion. These modules effectively improve the reconstruction quality, contributing to deal with very loose cloth and wide body motion.

\begin{figure*}[!htb]
\centering
\includegraphics[width=1\textwidth,trim={3cm 8.5cm 1cm 5.5cm},clip]{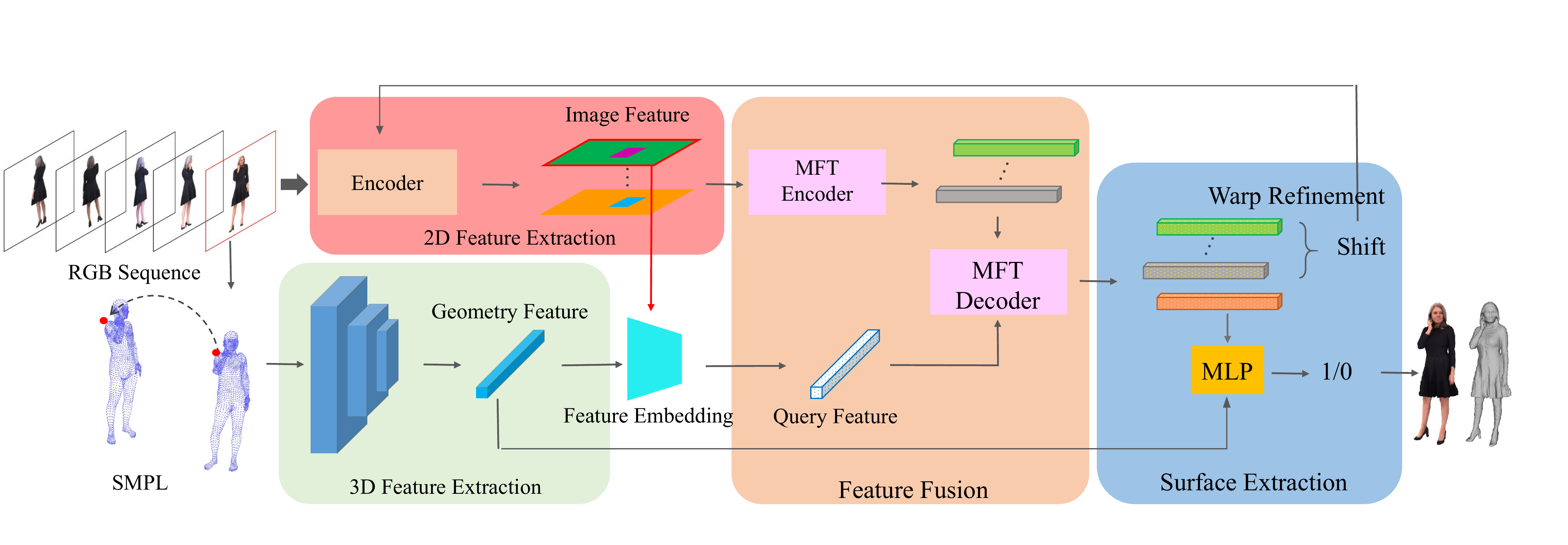}
\caption{Overview of our CrossHuman pipeline. The framework consists of five components: i) reference frames selection, SMPL-guided tracking and warping ii) image feature and SMPL geometry feature extraction iii) multi-frame feature fusion, iv) warp refinement v) occupancy or color inference. }
\label{img2}
\end{figure*}
%\section{Related Work}
\section{Related Work}

\subsection{Implicit Representations}

While explicit representations such as point cloud \cite{fan2017point, lin2018learning, qi2017pointnet, aliev2020neural}, voxels \cite{girdhar2016learning, yan2016perspective, sitzmann2019deepvoxels, lombardi2019neural} and triangular meshes \cite{alldieck2019tex2shape, lazova2019360, osman2020star, kolotouros2019learning, ma2020learning} have been explored to represent human 3D surface, these representations fail to represent high-quality 3D surfaces required for detailed human reconstruction.

Implicit 3D surface reconstruction methods \cite{saito2019pifu, saito2020pifuhd, chibane2020implicit, he2020geo} defines a surface as a level set of a function of occupancy probability or surface distance. 
Recent works have shown promising results on human shape reconstruction based on deep implicit representations. 
PIFu\cite{saito2019pifu} propose to regress a pixel-aligned implicit function and is able to reconstruct high-fidelity human surface based on pixel-level image features.
PIFuHD\cite{saito2020pifuhd} extend PIFu for capture detailed reconstruction results with additional fine-level feature extraction network.
Both methods lack the constrained priors of human body, which results in unstable performance especially for rare-seen poses.

\subsection{Deep Implicit Reconstruction based on Human Priors}

Previous works try to introduce human priors into deep implicit reconstruction. 
ARCH\cite{huang2020arch} and IP-Net\cite{bhatnagar2020combining} introduce SMPL model as parametric human representations. ARCH\cite{huang2020arch} proposed to regress 3D avatars in a canonical pose, but fails to generate accurate results, especially for
loose clothes.
IP-Net\cite{bhatnagar2020combining} jointly predict the 3D human model surface, the inner surface and body part labels from human point clouds.
PAMIR\cite{zheng2021pamir} conbines image features and SMPL volume features to reconstruct human surfaces, which can recover high-fidelity details in visible regions, but still over-smoothed in invisible areas in case of a single image setting. The SMPL volume feature helps PaMIR handle challenging poses, but when input SMPL is not accurate enough the reconstruction could be broken. 

\subsection{Tracking-based Human Reconstruction}

Tracking-based methods \cite{habermann2019livecap, habermann2020deepcap, newcombe2015dynamicfusion, yu2018doublefusion, yu2017bodyfusion, zhi2020texmesh} utilize a pre-scanned human model as template and infer the deformation. 
DoubleFusion\cite{yu2018doublefusion} propose a real-time human 3D reconstrcution system that combines volumetric reconstruction with parametric human model from a single depth camera. However, this method cannot effectively give the dynamic 3D reconstruction results of human body wearing very loose cloth and cannot effectively deal with the geometrical topological changes of human body.
POSEFusion\cite{li2021posefusion} leverages tracking-based and tracking-free methods to achieve high-fidelity human surfaces using a RGBD camera. It fuses the features extracted from keyframe, and obtain the information of the invisible area of the human body. However, POSEFusion cannot deal with very loose cloth and the hardware costs limit its popularity.

\section{Method}
An overview of our approach is illustrated in Fig.~\ref{img2}, the cross guidance learned from RGB sequence and parametric body contribute to the high-fidelity reconstruction.

We obtain the SMPL model from the whole RGB sequence, and select reference frames for each target frames (Section 3.1). In the training stage, we sample points near the 3D ground truth with similar strategies presented in PIFu. In the testing stage, we allocate a volume space and create a grid by resolution. We bind the points (voxels) to the SMPL vertices (Section 3.2) and warp the points (voxels) to reference frame spaces (Fig.~\ref{warp}), so that we can fuse features from different frames(Section 3.3). Fused features are used to predict occupancy or color (Section 3.5) through Multilayer Perceptron (MLP). The mesh can be extracted using Marching Cubes~\cite{lorensen1987marching}. However, warped points (voxels) are not necessarily close to the true human body surface when people wear loose cloth or the SMPL model is inaccurate. To this end, we propose a self-supervised training scheme for warp refinement (Section 3.4), which could correct the wrong warping and bring the warped point closer to the true human body surface.

\subsection{Reference Frames Selection}

From the whole RGB sequence, the current \textit{i}-th frame is described as target frame in this paper. Inspired by POSEFusion \cite{li2021posefusion}, we use pose-guided frame selection to choose reference frames. The key points for selecting reference frames are the complementarity of visible areas and the similarity of poses from other frames. We track the SMPL model from the whole RGB sequence, and select N (N = 4 in our experiments) reference frames.

According to the visibility complementarity, we can select reference frames from different angles of the human body to get a complete surface information about the human body. Pose similarity is defined as the similarity between the reference frame pose and the target frame pose. Based on pose similarity of the SMPL model, we calculate the pose-guided frame selection function between target frame and \textit{j}-th reference frame as:
\begin{equation}
 F_{pose} \left ( i, j \right ) = \sum_{k\in \mathcal{J}}\omega _{k} \left |\theta  _{i}^{k} -\theta  _{j}^{k} \right |^{2}   + \frac{\omega _{{k}' }}{\left |  \theta  _{i}^{{k}' }-\theta  _{j}^{{k}' }\right | ^{2}}
\end{equation}
where $\mathcal{J}$ is the joint index set of the SMPL model expcet the global rotation and translation, ${k}'$ is the global rotation and translation, and the $\omega_{k}$ is the weight of the \textit{k}-th joint to the reference frame, $\theta$ denotes the rotation angle for the joint. In this work, the whole pipeline is assumed to be off-line, so we can precompute reference frames for each frame in a video sequence before running reconstruction frame by frame.

\subsection{SMPL-guided Tracking}
\textbf{Parametric body Model}
Inspired by POSEFusion and PaMIR, we integrate the parametric SMPL model to guide the warp of space and regularize the human reconstruction. SMPL is a skinned model with 6890 vertices, and it is defined as a function of shape parameters and pose parameters. The function $M\left ( \theta,\beta   \right ) $ is formulated as:
\begin{equation}
\begin{aligned}
M\left ( \theta,\beta   \right ) &= W(T(\theta,\beta), J(\beta), \theta, \mathcal{W}) \\
T(\beta, \theta ) &= \bar{\mathrm{T} } +B_{\mathit{s} } (\beta )+B_{\mathit{p} } (\theta )
\end{aligned}
\end{equation}
where W($\cdot$) is a linear blend-skinning function with deformed body shape T($\cdot$), pose parameters $\theta$, skeleton joints $J(\beta)$ and skinning weights $\mathcal{W}$, which returns the posed vertices. The deformed SMPL model can also serve as a geometry prior, which regularize the human reconstruction.
\begin{figure}[h]
\centering
\includegraphics[width=0.48\textwidth,trim={3.6cm 0.2cm 0cm 0cm}, clip]{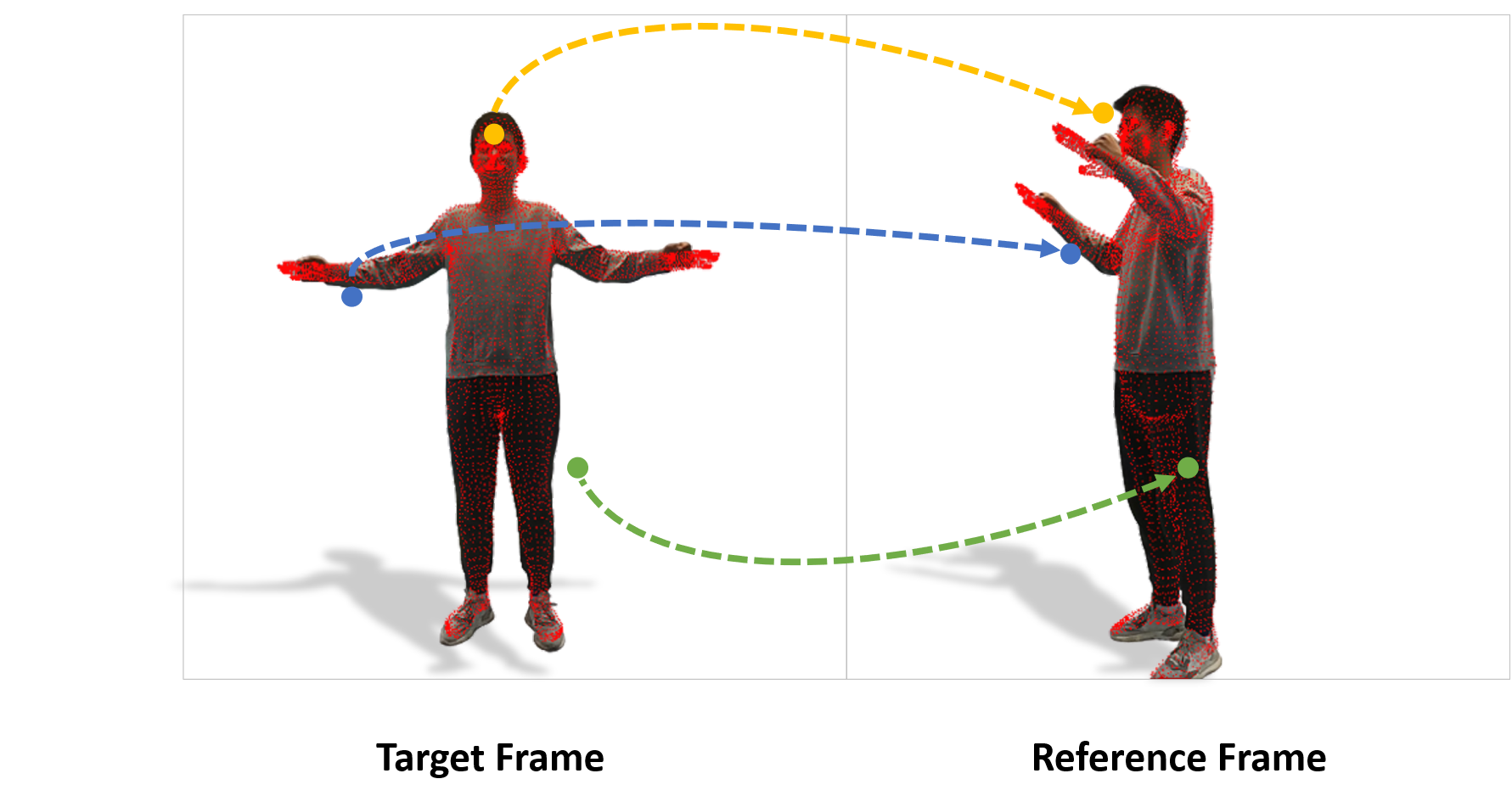}
\caption{Points Warping. Points bound to the SMPL model are warped to reference frame spaces to fetch corresponding features through body motion tracking.}
\label{warp}
\end{figure}

\noindent\textbf{Sampling, Binding and Warping}
To ensure the temporal continuity, we use video based SMPL estimation method \cite{kocabas2020vibe} to track the SMPL model from the whole sequence. The SMPL estimation module used in our framework can be replaced by any other SMPL estimation algorithm for monocular video . According to PIFu \cite{saito2019pifu}, the sampling strategy can affect the quality of reconstruction. In the training stage, we first randomly sample points on ground truth body surface in the target frame space and add offsets with normal distribution N(0, $\sigma$) ($\sigma$ = 0.05 cm in our experiments) for x, y, and z axis. We then uniformly samples points within a 3D bounding box containing the ground truth body. Finally, We mix uniformly samples points and points nearby the surface. In the testing stage, we allocate a volume space and create a grid by resolution. After sampling, we bind the points (voxels) to SMPL vertices and warp the points (voxels)  to reference frame space by SMPL motion(Fig.~\ref{warp}). The warp matrix $M_{i} (\theta ,\beta)$ of $v_{\textit{i}}$ from T-pose to the other pose is computed by:

\begin{equation}
\mathbf{M} _{i} (\theta ,\beta)=\left ( \sum_{j=1}^{K}  W _{i, j}\mathbf{G} _{j}\right )
\left[
  \begin{array}{cc}
    \mathbf{I} & \mathbf{B} _{S, i}(\beta )+\mathbf{B} _{P, i}(\theta ) \\
    \mathbf{0} ^{T} & 1\\
  \end{array}
\right]
\end{equation}
where the blend weight W$_{i, j}$ is defined as the influence of rotation of part \textit{j} on the vertex $v_{\textit{i}}$, $\mathbf{G} _{j}$ is the world transformation of joint \textit{j}, $\mathbf{B} _{S, i}(\beta )$ is pose blendshape, and $\mathbf{B} _{P, i}(\theta )$ is shape blendshape. 

we define the warp of SMPL vertices from one pose $x_{1}$ to another pose $x_{2}$ in the SMPL coordinate system as:
\begin{equation}
\left[
  \begin{array}{c}
    \mathbf{x_{2}}  \\
    1 \\
  \end{array}
\right]=
\mathbf{M_{\mathit{i} }} (\beta, \theta_{2})\mathbf{M_{\mathit{i} }} ^{-1}(\beta, \theta_{1})
\left[
  \begin{array}{c}
    \mathbf{x_{1}}  \\
    1 \\
  \end{array}
\right]
\end{equation}

\begin{figure}[h]
\centering
\includegraphics[width=0.32\textwidth,trim={0.5cm 0.5cm 0cm 0cm}, clip]{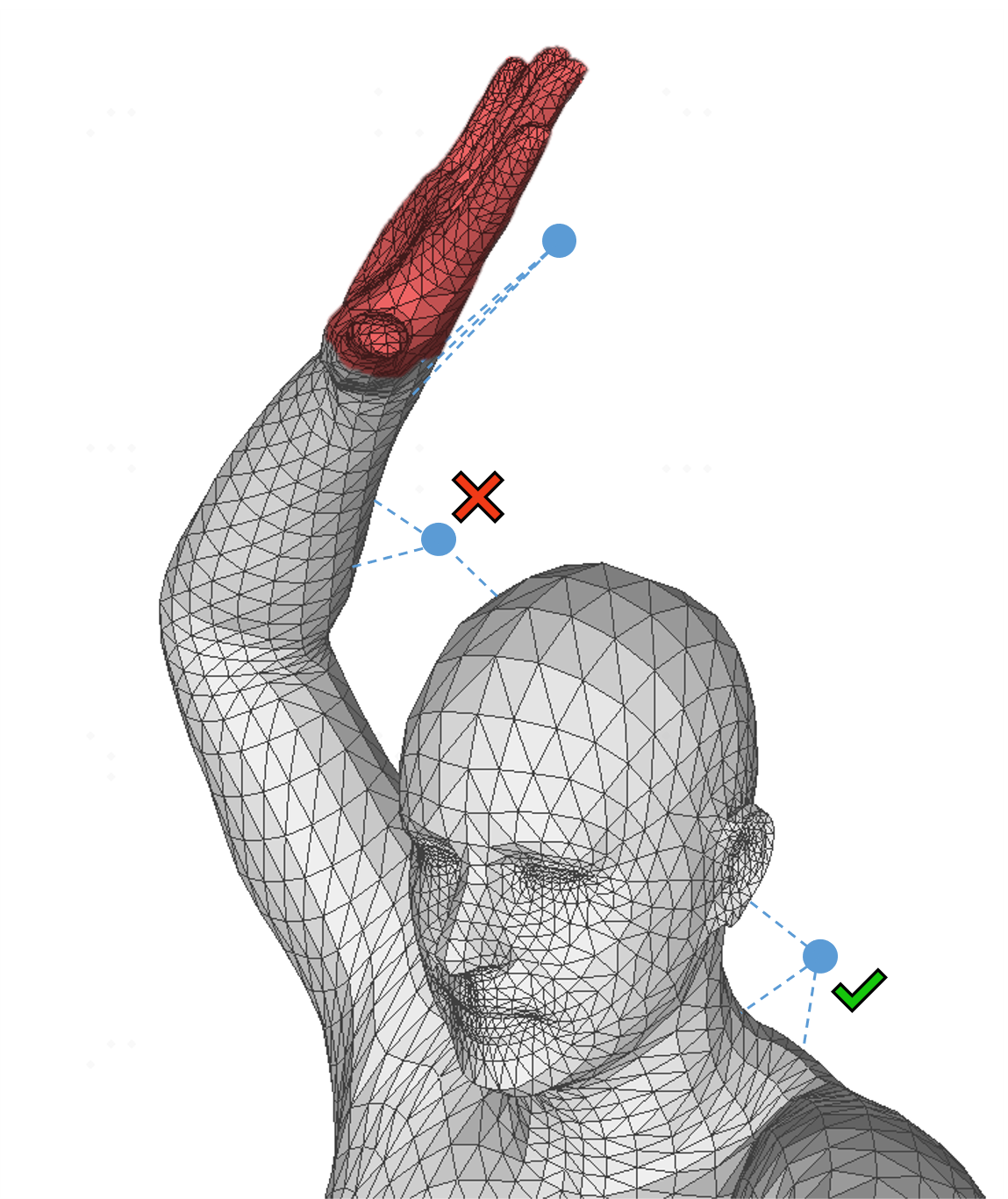}
\caption{Points Binding}
\label{bind}
\end{figure}

We found that binding strategies influence the accuracy of warp. Consider the fact that a sampled point might be bound to different body parts, leading to non-meaningful or even wrong warp. In order to improve the accuracy of warp, we bind each sampled point to the three SMPL vertices closest to it, and divide the SMPL vertices into 14 parts, including hand, head and torso, etc. If the SMPL vertices bound to one sampled point belong to the same part or neighboring parts, the warp matrix of the three SMPL vertices will be weighted. If the bound SMPL points belong to the non-neighboring parts, the sampled point will be ignored (Fig.~\ref{bind}). The warp weight is designed as:
\begin{equation}
\begin{aligned}
\omega _{j\longrightarrow i}&=exp(\frac{\left \|  p_{j}-v_{i}\right \| }{2\sigma ^{2}} ) \\
\omega _{i}&=\sum_{j\in \mathcal{N}(i)}\omega _{j\longrightarrow i}
\end{aligned}
\end{equation}
Where $p_{\textit{j}}$ is a 3D sampled point, $v_{\textit{i}}$ is a SMPL vertex, $\mathcal{N}(i)$ is the SMPL vertex closest to the $p_{\textit{j}}$ and $\mathcal{N}(i)=3$, $\omega _{j\longrightarrow i}$ is the binding weight, and $\omega _{i}$ is the weight normalizer.

We can warp the  sampled points from one pose to another pose as:
\begin{equation}
\left[
  \begin{array}{c}
    \mathbf{p_{2}}  \\
    1 \\
  \end{array}
\right]=
\sum_{j\in \mathcal{N}(i)}
\frac{\omega_{j\longrightarrow i} }{\omega _{i}} 
\mathbf{M_{\mathit{i} }} (\beta, \theta_{2})\mathbf{M_{\mathit{i} }} ^{-1}(\beta, \theta_{1})
\left[
  \begin{array}{c}
    \mathbf{p_{1}}  \\
    1 \\
  \end{array}
\right]
\end{equation}
\\Due to the limitation of the SMPL parametric model, the estimated hands and feet of the SMPL body are usually not accurate, which affects the warp of the hand and foot. In this work, we shielded the hands and feet from binding (Fig.~\ref{bind}).

\begin{figure*}[h]
\centering
\includegraphics[width=1\textwidth,trim={0cm 4.9cm 6cm 2cm}, clip]{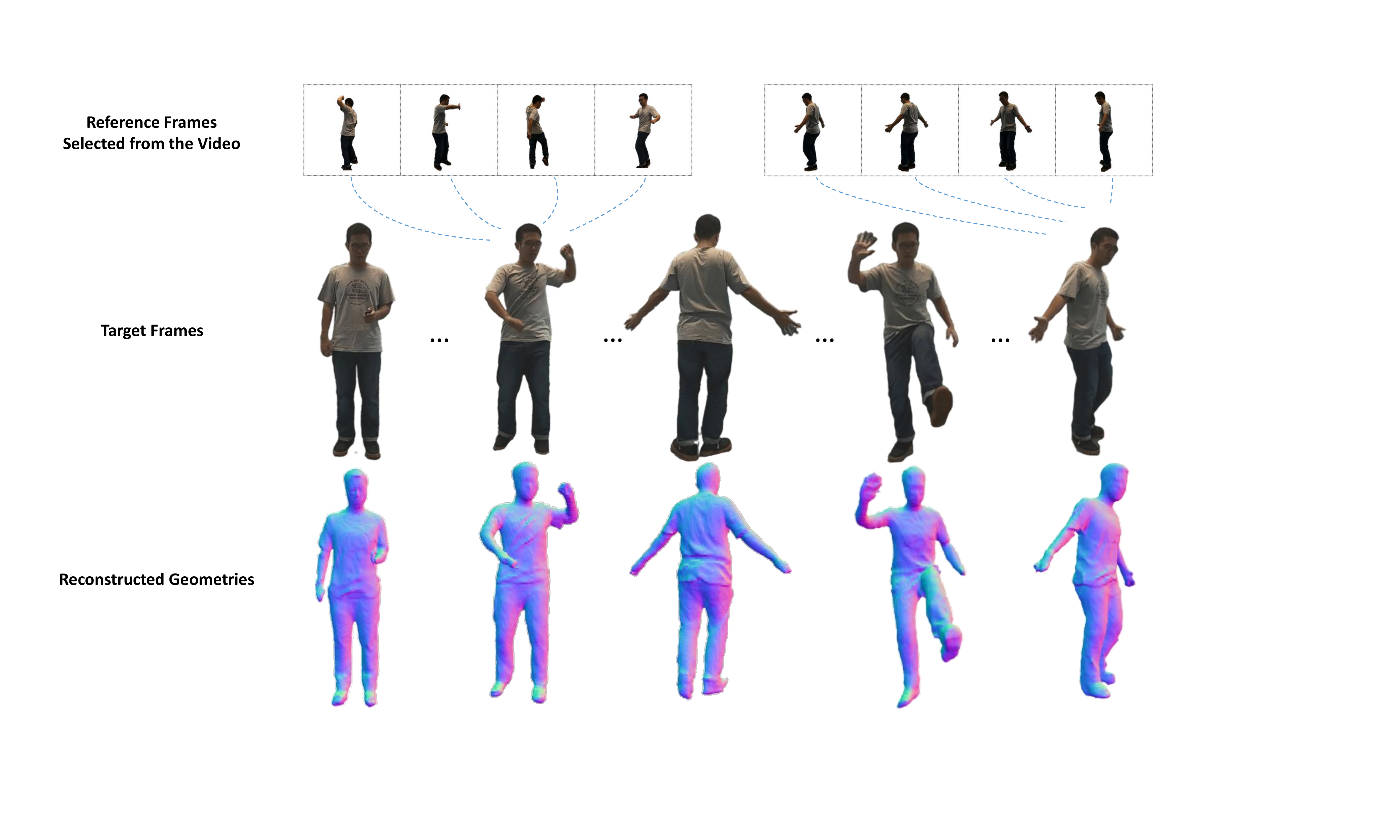}
\caption{Dynamic Results on Real Video}
\label{img3}
\end{figure*}

\subsection{Feature Fusion}
% \textbf{deep implicit function}
% Our method is base on the SMPL model and deep implicit function. A deep implicit function defines the surface of a 3D model as the level set of an occupancy field function F, e.g. F(X) = 0.5, where X is a 3D point. Specifically, In PIFu, the function F also takes a conditional variables as input, and can be formulated as:
% \begin{equation}
% F(\Phi(\mathbf{x} , \mathbf{I} ),z(\mathbf{X} ))=s:s\in [0,1]
% \end{equation}
% where $\mathbf{I}$ is the input image, $\mathbf{X}$ is a given 3D point, $\pi(X)$ is the 2D projection of x, $\Phi(\mathbf{x} , \mathbf{I} )$ is the image embedding feature, and z($\mathbf{X}$) is the depth value in the camera space.

\textbf
Thanks to the SMPL-guided tracking, one can achieve spatially aligned local features by projecting warped points to multi-frame feature maps. Straight-forward ways to fuse those features from different frames are operations like average pooling \cite{saito2019pifu} or self-attention \cite{zheng2021deepmulticap}. To force SMPL geometry and multi-frame RGB jointly guide the feature fusion, we design a multi-frame transformer(MFT). The MFT adopts self-attention based enocder-deocder architecture inspired by view-to-view transformer proposed in DoubleField \cite{shao2021doublefield}.  

The MFT encoder fuse features from multi-frame images with self-attention operation in Eq[7]:

\begin{equation}
\begin{aligned}
\varphi^{i} &= \varphi(x, I_{i}) \\
Q^{e}, K^{e}, V^{e}&=F_{Q, K, V}^{e} (\varphi^{1},\varphi^{2},...,\varphi^{n} )\\
\Phi &=F^{e}(Att(Q^{e}, K^{e}, V^{e}))
\end{aligned}
\end{equation}
where $\varphi^{i}$ is pixel-aligned feature of points x on image $I_{i}$. $F_{Q, K, V}^{e}(\cdot)$ denotes the linear layers, which output the query, key and value matrices, Att($\cdot$) is the multi-head attention operation, $F^{e}$ is the feed-forward layer.

A feature embedding combined with target frame image feature and SMPL geometry feature works as the query feature in MFT deocder. The decoder enables the cross-attention between the query feature and multi-frame features. To be more specific, the operation can be formulated as:
\begin{equation}
\begin{aligned}
\phi_{q}&=E_{q}(\oplus (\varphi(x, I_{t}),\Psi (x,S_{t}))) \\
\phi  _{Q}, \phi _{K}, \phi _{V} &= F_{Q, K, V}(\Phi,\phi_{q}) \\
e_{o}, e_{s} &=F^{e}(att(\phi  _{Q}, \phi _{K}, \phi _{V}))
\end{aligned}
\end{equation}
where $\Psi(\cdot)$ is a function extracting SMPL geometry feature. $S_{t}$ is the SMPL parameters for target frame, $\oplus$ is a concatenate operator, $E_{q}$ is used to generate query feature embedding, $F_{Q, K, V}$ is the linear layers, att($\cdot$) is the attention operation in the transformer, and $F^{e}$ is  the feed-forward layer.

The MFT decoder outputs a fused feature embedding $e_{o}$ and a multi-frame feature block $e_{s}$. The fused feature embedding is used for occupancy and texture inference through MLP $g$ and the multi-frame feature block is fed to another MLP $s$ for predicting warp refinement.   

\begin{equation}
g(\oplus(e_{o},\Psi (x,S_{t}))) = Occupancy 
\end{equation}

\begin{equation}
s(e_{s}) = Noise
\end{equation}

% To obtain complete human 3D reconstruction, we introduce CrossHuman. In CrossHuman, the implicit surface is fusion guided by the geometry features of the SMPL model and pixel aligned features from selected images. Instead of simply average the multi-view feature embeddings from MLP [PIFu], we propose a multi-view feature fusion method based on self-attention mechanism, and we define it as:
% \begin{equation}
% attention(\phi_{q}, \phi _{s}, \phi_{t})=softmax(\frac{\phi_{q}^{T} }{\sqrt{d_{k}} } )\phi_{t}
% \end{equation}
% \begin{equation}
% \phi_{f}=\Psi(\Phi (\pi(\mathbf{X}),\mathbf{I}),\varphi (\mathbf{X}, S_{t}))
% \end{equation}
% where $\phi_{q}=\phi_{f}\mathbf{W}_{q}$, $\phi_{s}=\phi_{f}\mathbf{W}_{s}$, and  $\phi_{t}=\phi_{f}\mathbf{W}_{t}$ are defined  as query, sxource and target feature embedded by weights $\mathbf{W}$, $\Psi (\cdot)$ is a function that stack pixel aligned features $\Phi (\cdot)$ and the geometry features $\varphi (\cdot)$, and $S_{t}$ is the SMPL parameters. 
% Finally, the meta-view prediction is denoted as:
% \begin{equation}
% C(X)=g(A(\phi _{f}))
% \end{equation}
% where $A(\phi _{f})$ is the fuesd feature output of the self-attention encoder, and the deep implicit function g($\cdot$) represented by a multi-layer perceptron. It predicts a continuous occupancy field of any 3D query point.

\subsection{Warp Refinement}
For the network that utilizes parametric human model to reconstruct human body, the accuracy of the parametric human model greatly affects the accuracy of the reconstruction system. In this work, we propose a self-supervised training scheme, which could improve the accuracy of the reconstruction system in the cases of estimated inaccurate SMPL model and very loose cloth. In training stage, we randomly add noise shift to warped 3D points and feed the multi-frame feature block output by MFT to an MLP for predicting the random noise shift. We adopt iterative error feed back strategy \cite{kanazawa2018end,zhang2021pymaf} for progressive shift prediction. We jointly minimize the Self-supervised shift loss $L_{s}$ and occupancy loss $L_{o}$:
\begin{equation}
\mathcal{L} _{s}=\frac{1}{n}\sum_{ief=1}^{t} \sum_{i=1}^{n} \left |  s_{i}-s^{*}_{i}\right | 
\end{equation}
\begin{equation}
\mathcal{L} _{o}=\frac{1}{n}\sum_{i=1}^{n} \left |  g_{i}-g^{*}_{i} \right | 
\end{equation}
\begin{equation}
\mathcal{L}=\mathcal{L} _{s}+\mathcal{L} _{o}
\end{equation}
where $s_{i}$ is the predicted shift for point $X_{i} \in \mathbb{R} ^{3}$,and $s^{*}_{i}$ is the added noise. $g_{i}$ is the predicted occupancy,and $g^{*}_{i}$ is the ground truth occupancy. Our ablation study demonstrate such training strategy contributes to higher reconstruction accuracy. 
\subsection{Texture Inference}

After recovering the mesh of the target frame, we warp reconstructed vertices to each reference frames to fetch image feature extracted by another image encoder. We also concatenate this feature with the geometry feature and than use MLP to predict the vertex color, which is similar to the process of occupancy prediction.

\begin{table*}[]
\centering
\caption{Comparison of our method with other state-of-the-art works.}
\begin{tabular}{llllllll}
\toprule
Methods & Input & \begin{tabular}[c]{@{}l@{}}Topological\\ Change\end{tabular} & \begin{tabular}[c]{@{}l@{}}Natural\\ Deformation\end{tabular} & \begin{tabular}[c]{@{}l@{}}Details in\\ Invisible Regions\end{tabular} & Texture   & Loose Cloth \\
\midrule
DoubleFusion\cite{yu2018doublefusion} & Monocular Depth & $\times$ & $\times$ & $\surd$ &  None & $\times$          \\
TexMesh\cite{alldieck2019tex2shape}      & Monocular RGBD & $\times$ & $\times$ & $\surd$ &  $\surd$ & $\times$     \\
PIFu/PIFuHD\cite{saito2019pifu,saito2020pifuhd}  & Monocular RGB   & $\surd$ & $\surd$ & $\times$  & $\surd$ & $\surd$ \\
Multi-view PIFu/PIFuHD\cite{saito2019pifu,saito2020pifuhd}  & Multiple Calibrated RGB   & $\surd$ & $\surd$ & $\surd$  & $\surd$ & $\surd$ \\
PaMIR\cite{zheng2021pamir}        & Monocular RGB   & $\surd$ & $\surd$ & $\times$ &  $\surd$ & $\surd$  \\
Multi-frame PaMIR\cite{zheng2021pamir}        & Monocular RGB   & $\surd$ & $\surd$ & $\surd$(limited) &  $\surd$ & $\times$  \\
POSEFusion\cite{li2021posefusion}   & Monocular RGBD & $\surd$ & $\surd$ & $\surd$ &  $\surd$  & $\surd$(limited)  \\
OURS         & Monocular RGB  & $\surd$ & $\surd$ & $\surd$  &  $\surd$ & $\surd$\\  
\bottomrule

\end{tabular}
\label{mark}
\end{table*}

\section{Experiments}
\subsection{Datasets}

Our goal is to reconstruct complete human model from RGB sequence, but there are few high-quality 4D public datasets. To solve it, We collect 1123 and 317 human scans from RenderPeople \cite{renderpeople} dataset and THuman2.0 \cite{yu2021function4d} dataset respectively, which contain high-quality textured human meshes, complex clothing (e.g. long skirts and down jackets) and challenging poses, then we render the scans in these two datasets using a perspective camera model, and rotate the human model for 360 degrees in yaw axis. We split our RenderPeople dataset into a training set of 1040
subjects and a testing set of 83 subjects. Our model for evaluation are trained with our Renderpeople training set. THuman2.0 dataset and RenderPeople testing set are used for evaluation.

\begin{figure*}[]
\centering
\includegraphics[width=0.98\textwidth,trim={5.1cm 1.6cm 1.3cm 2.7cm}, clip]{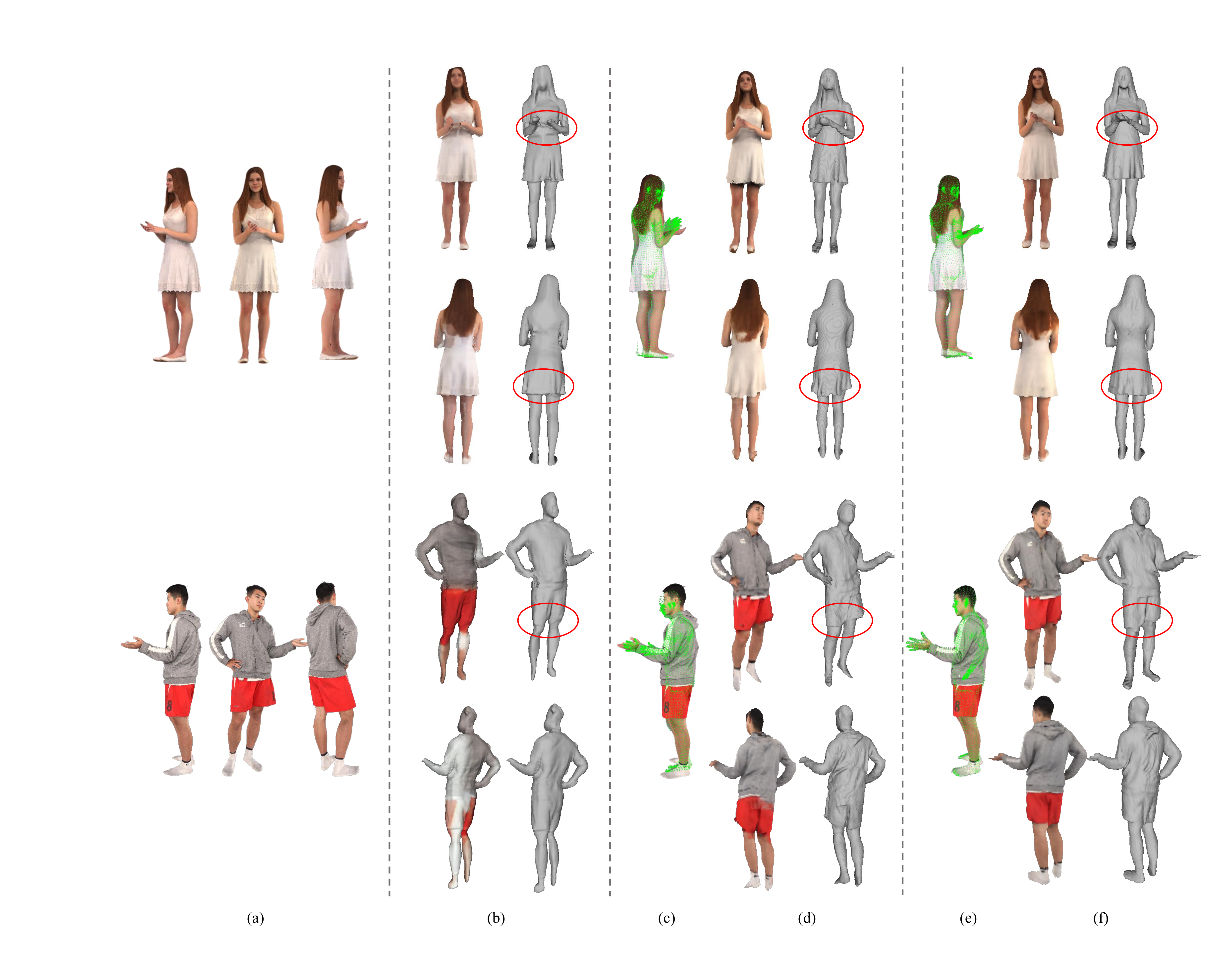}
\caption{Qualitative comparison with SOTA methods (a) three images of the input sequence (b) Multi-Frame PaMIR (c) estimated SMPL vertices projected on image (d) ours with estimated SMPL (e) GT SMPL vertices projected on image (f) ours with GT SMPL.}
\label{img4}
\end{figure*}

% \newpage
\subsection{Implementation Details}

For multi-frame image feature extraction, we choose HR-Net \cite{sun2019deep} as the image encoder, which outputs a 256-channel feature map. For geometry feature extraction, inspired by \cite{peng2021neural, peng2020convolutional, shi2020pv}, we choose the SparseConvNet \cite{graham20183d} to process the SMPL model, divide the 3D bounding box of the SMPL into voxels and obtain a 352-channel geometry feature volume. We adopt a multi-frame transformer for feature fusion and feed the fused feature to a MLP to predict occupancy, where the number of neurons in this MLP is (608, 1024, 512, 256, 128, 1). For the texture network, we extract multi-frame image feature and stack it with previous geometry feature volume. The decoder is implemented as a MLP, where the number of neurons is (864, 1024, 512, 256, 128, 3). The number of neurons in the MLP for noise shift prediction is (256, 512, 256, 128, 3). The artificially added noise shift is generated by Gaussian distribution. 

To evaluate our CrossHuman, we use multi-frame images as input, which are rendered using a perspective camera model or a weak-perspective camera model, and we selected images of challenging poses or complex clothing as input to demonstrate the robustness of our network. What is more, we use real world data to demonstrate the generalization ability of our network.

% \newpage

\section{Evaluations}

\subsection{Results on Real Video}
We demonstrate our method for dynamic high-quality 3D human reconstruction in Fig.~\ref{img3}. The input is a real RGB video, and human poses are continuously changed in frames. The results demonstrate our CrossHuman trained with static synthetic data without nonrigid pose change among frames can generalize well on real video data and achieve high quality dynamic results. The video results can be seen in the supplementary material.

% can reconstruct high-fidelity geometry and texture in both visible and invisible regions and can handle challenging poses and loose cloth. More importantly, our network has good generalization.
\begin{table}[]
\caption{Quantitative results on RenderPeople and THuman2.0 dataset.}
\begin{threeparttable}
\begin{tabular}{lcc|cc}
\hline
                             & \multicolumn{2}{c|}{Renderpeople} & \multicolumn{2}{c}{THuman2.0} \\
                             & Chamfer$\downarrow$         & P2S$\downarrow$           & Chamfer$\downarrow$  & P2S$\downarrow$         \\ \hline
Octopus\cite{alldieck2019learning}                      & 5.77                         & 5.63                       & 5.84                          &  6.02         \\
Multi-view PIFu\cite{saito2019pifu}*             & 0.77                      & 0.79                    & 1.88                       & 2.16 \\
Multi-frame PaMIR\cite{zheng2021pamir}            & 1.90                         & 2.33                       & 5.37                          & 5.11          \\
Ours w/ estimated SMPL       & 0.75                         & 0.81                       & 2.14                          & 2.32          \\\hline 
Ours w/ GT SMPL\dag              & 0.63                     &0.76                       &1.80                          & 2.02          \\ \hline 
\end{tabular}

\begin{tablenotes}
        \footnotesize
        \item* Multi-view PIFu requires well-calibrated multi-view input, other methods only require monocular RGB sequence. ${\dag}$ Our results with GT SMPL indicates the upper bound of our method. %此处加入注释*信息
      \end{tablenotes}
    \end{threeparttable}
    
\label{quant}
\end{table}

% \noindent \textbf{Comparisons}
\subsection{Comparisons}
As is shown in Table.~\ref{mark}, our method inherits all the advantages of other SOTA human reconstruction methods while avoiding their drawbacks. Moreover, our method can be reproduced with inexpensive hardware and convenient settings. In Fig.~\ref{img4}, We qualitatively compare our approach  with Multi-frame PaMIR \cite{zheng2021pamir} which is a state-of-the-art method to recover 3D human from multi frame images. Besides, We present the results of our method under estimated SMPL and ground truth SMPL (We register SMPL to scans to obtain ground truth SMPL). It is obvious that wide topological changes like the fluttering gym shorts give PaMIR a lot of trouble, while CrossHuman with MFT and warp refinement is able to handle natural deformation and topological change. The results prove our CrossHuman can recover high-fidelity details under challenging poses and loose clothes. Meanwhile, our method can ensure good performance even under imperfect SMPL. The visualization on column (c) shows the estimated SMPL vertices cannot align well with the target human on image plane, including hands region, feet region and face. CrossHuman can still generate plausible and detailed results. Such SMPL fault tolerance makes our method applicable in a wider range of conditions In addition, we conduct quantitative comparison in Table.~\ref{quant}. We compute point-to-surface Euclidean distance (P2S) in cm from the vertices on the reconstructed mesh to the ground truth. We also compute Chamfer distance in cm from the reconstructed mesh and the ground truth mesh. Results in the Table.~\ref{quant} demonstrate our method improves the reconstruction accuracy. The multi-view PIFu listed in Table.~\ref{quant} requires well calibrated cameras.

% \noindent \textbf{Ablation study}
\subsection{Ablation study}
\begin{figure}[]
\centering
\includegraphics[width=0.5\textwidth,trim={9cm 2.8cm 9cm 1cm},clip]{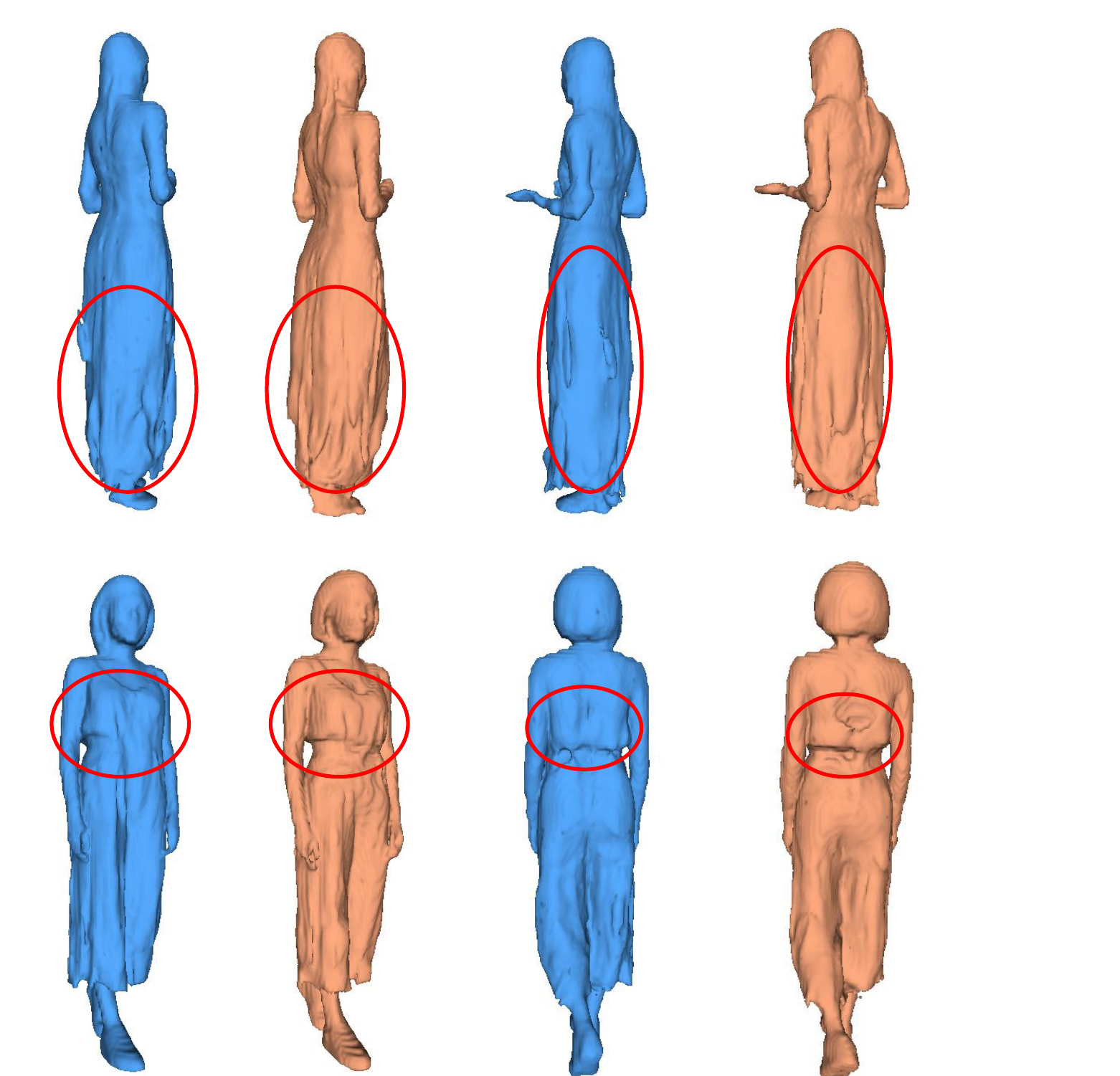}
\caption{Qualitative evaluation on feature fusion module. The blue models are our reconstructions with self-attention block, and the orange models are our reconstructions with MFT.}
\label{img5}
\end{figure}

\begin{figure}[]
\centering
\includegraphics[scale=0.5,trim={0cm 2.47cm 20cm 0cm},clip]{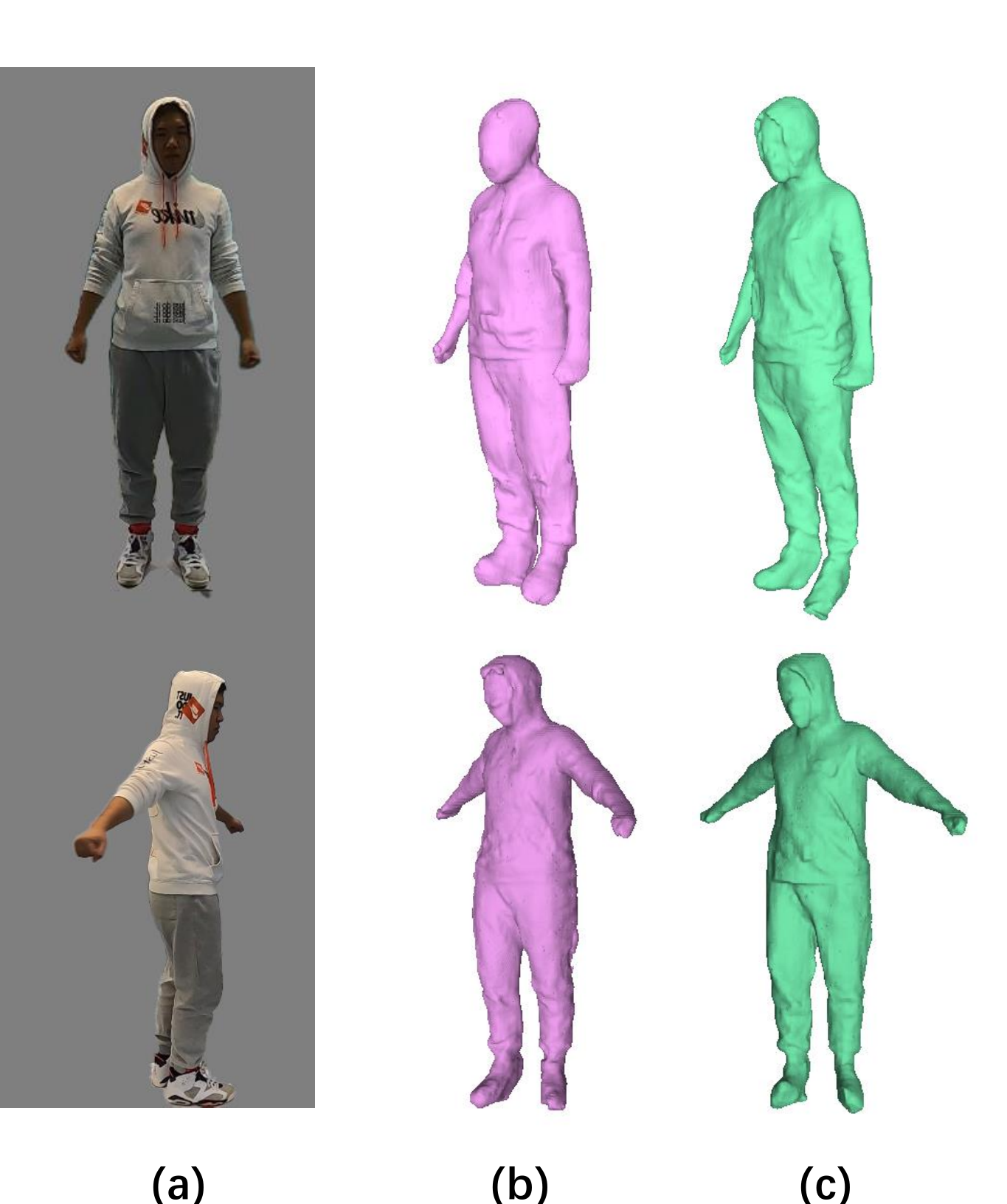}
\caption{Qualitative evaluation of warp refinement module on real video (a) two frames of input sequence (b) our reconstruction results wo/ warp refinement (c) our reconstruction results w/ warp refinement.}
\label{img6}
\end{figure}

\begin{figure}[]
\centering
\includegraphics[scale=0.57,trim={0cm 0cm 0cm 0cm},clip]{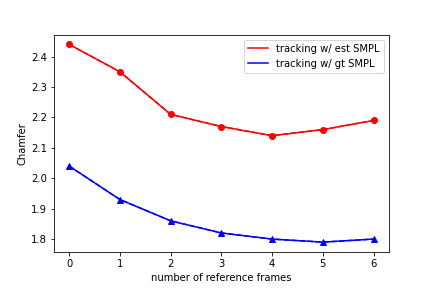}
\caption{Evaluation of the number of reference frames}
\label{track}
\end{figure}

% Inspired by DeepMultiCap [], We use a V2V-Transformer to capture the information from multi-frame images and SMPL model. In order to prove the effectiveness of our method in fusing multi-frame image features and geometric features, we use the multi-view feature fusion method based on average pool to replace our fusion method. The results in Figure 5 show that our method can effectively recover more surface detail in both visible and invisible regions. To verify the effect of self-supervised  training  scheme, we conducted several comparative experiments. Fig. 6 illustrates the results of our method with and without self-supervised training scheme under inaccurate SMPL estimation and the results of different iterations. The self-supervised training scheme improves the robustness of the human reconstruction system under challenging poses and loose cloth. What is more, we design a deform loss function, and train the CrossHuman by minimizing Self-supervised  deform  loss and occupancy loss. From Fig. 7, we can see that the convergent rate of the network is improved by inducting the deform loss.
To figure out the strength of our multi-frame transformer, we train another two models with average pooling feature fusion used in PIFu\cite{saito2019pifu} and self-attention based feature fusion strategy used in PixelNerf \cite{yu2021pixelnerf} and DeepMultiCap \cite{zheng2021deepmulticap}. The results in Fig.~\ref{img5} demonstrate that our method with MFT can effectively recover more surface detail in both visible and invisible regions. In addition, we conduct comparative experiments to verify the effect of the self-supervised warp refinement training module. Fig.~\ref{img6} illustrates the results of our method on real video data with and without warp refinement module under inaccurate SMPL estimation, which prove the warp refinement training module can effectively reduce some artifacts and blurring caused by tracking errors. Owing to these two modules, CrossHuman is able to generate visually coherent and robust reconstruction results in most cases including cases with loose cloth and challenging poses. Results in Table.~\ref{quant_eva} gives quantitative evidence of the effectiveness of our MFT and warp refinement scheme. We also conduct an ablation experiment on the THuman dataset for tuning the appropriate number of reference frames. As is shown in Fig.~\ref{track}, the larger number of reference frames doesn't necessarily lead to better performance, since the multi-frame feature fusion will produce a smoothing effect.

\begin{table}[]
\caption{Quantitative evaluations on Renderpeople dataset with estimated SMPL.}

\begin{tabular}{lcl}
\hline
\multicolumn{1}{c}{}           & Chamfer$\downarrow$       & P2S$\downarrow$           \\ \hline
average pooling                & 1.11          & 1.73          \\
self-attention                 & 0.96          & 1.02          \\
ours (MFT wo/ warp refinement) & 0.83          & 0.90          \\
ours (MFT w/ warp refinement)  & 0.75          & 0.81           \\ \hline
\end{tabular}%
\label{quant_eva}

\end{table}

% \begin{table}[h]
% \caption{Quantitative results on several datasets}
% \centering
% \begin{tabular}{lllll}

% & \multicolumn{3}{c}{Common Poses} & \multicolumn{2}{|c}{Challenging Poses}  \\
%                   & CD\downarrow    & P2S\downarrow     & Norm\uparrow    & CD\downarrow             & P2S\downarrow               \\ \hline
% \midrule
% Octopus
% Multi-frame PaMIR       & 1.90          & 2.33             & 5.37      & 5.11             \\
% Ours w/ estimated SMPL  & 0.75          & 0.81             & 2.14      & 2.32             \\
% Ours w/ GT SMPL         & 0.63          & 0.76             & 1.80      & 2.02     \\
% \bottomrule

% \end{table}

\subsection{Limitation and Discussion}
Many previous monocular human reconstruction methods combined with implicit function and parametric models did not obtain satisfactory robustness especially for challenging poses. A key difficulty is to ensure the perfect alignment between the parametric body and the human on the image, which is not easy on 2D image plane and even harder in 3D space due to depth ambiguity. From the point of view of the practicality of the algorithm, we sidestep this problem by regularizing the reconstructed geometry with parametric body strictly. To be more specific, our CrossHuman always trust the input parametric body no matter whether it is accurate or not and rely on MFT and warp refinement to deal with tracking error. As a result, we can always get plausible results without mutilated limb even under the rough alignment. This strategy is visually successful but still a stopgap, making the quantitative reconstruction accuracy subject to the SMPL accuracy. A possible more comprehensive solution could be an end-to-end method to jointly infer the parametric body and detailed surface with mandatory alignment. Besides, though  our CrossHuman trained on static data generalize well on real-video with non-rigid deformation, the details of invisible regions are not natural enough in some cases. A model trained with 4D dataset is expected to handle more natural non-rigid deformation.

\section{Conclusion}

In this paper, we propose CrossHuman, the method that learns cross-guidance from RGB sequence and corresponding parametric human model to achieve high-quality human 3D reconstruction from a single RGB camera. we propose a new self-supervised warp refinement scheme and a novel multi-frame transformer to relax the requirement of SMPL model accuracy, leading to the robust performance in case of very loose cloth even under imperfect SMPL estimation. The detailed dynamic reconstruction results on real video prove the good generalization of our method. We believe our work will contribute to the development of cost-effective 3D human reconstruction, providing more possibilities for consumer-grade VR/AR applications and immersive entertainment or communication.

\newpage

%%
%% The next two lines define the bibliography style to be used, and
%% the bibliography file.
\bibliographystyle{ACM-Reference-Format}
\bibliography{sample-base}

%%% -*-BibTeX-*-
%%% Do NOT edit. File created by BibTeX with style
%%% ACM-Reference-Format-Journals [18-Jan-2012].

\begin{thebibliography}{61}

%%% ====================================================================
%%% NOTE TO THE USER: you can override these defaults by providing
%%% customized versions of any of these macros before the \bibliography
%%% command.  Each of them MUST provide its own final punctuation,
%%% except for \shownote{}, \showDOI{}, and \showURL{}.  The latter two
%%% do not use final punctuation, in order to avoid confusing it with
%%% the Web address.
%%%
%%% To suppress output of a particular field, define its macro to expand
%%% to an empty string, or better, \unskip, like this:
%%%
%%% \newcommand{\showDOI}[1]{\unskip}   % LaTeX syntax
%%%
%%% \def \showDOI #1{\unskip}           % plain TeX syntax
%%%
%%% ====================================================================

\ifx \showCODEN    \undefined \def \showCODEN     #1{\unskip}     \fi
\ifx \showDOI      \undefined \def \showDOI       #1{#1}\fi
\ifx \showISBNx    \undefined \def \showISBNx     #1{\unskip}     \fi
\ifx \showISBNxiii \undefined \def \showISBNxiii  #1{\unskip}     \fi
\ifx \showISSN     \undefined \def \showISSN      #1{\unskip}     \fi
\ifx \showLCCN     \undefined \def \showLCCN      #1{\unskip}     \fi
\ifx \shownote     \undefined \def \shownote      #1{#1}          \fi
\ifx \showarticletitle \undefined \def \showarticletitle #1{#1}   \fi
\ifx \showURL      \undefined \def \showURL       {\relax}        \fi
% The following commands are used for tagged output and should be
% invisible to TeX
\providecommand\bibfield[2]{#2}
\providecommand\bibinfo[2]{#2}
\providecommand\natexlab[1]{#1}
\providecommand\showeprint[2][]{arXiv:#2}

\bibitem[\protect\citeauthoryear{Aliev, Sevastopolsky, Kolos, Ulyanov, and
  Lempitsky}{Aliev et~al\mbox{.}}{2020}]%
        {aliev2020neural}
\bibfield{author}{\bibinfo{person}{Kara-Ali Aliev}, \bibinfo{person}{Artem
  Sevastopolsky}, \bibinfo{person}{Maria Kolos}, \bibinfo{person}{Dmitry
  Ulyanov}, {and} \bibinfo{person}{Victor Lempitsky}.}
  \bibinfo{year}{2020}\natexlab{}.
\newblock \showarticletitle{Neural point-based graphics}. In
  \bibinfo{booktitle}{\emph{Computer Vision--ECCV 2020: 16th European
  Conference, Glasgow, UK, August 23--28, 2020, Proceedings, Part XXII 16}}.
  Springer, \bibinfo{pages}{696--712}.
\newblock


\bibitem[\protect\citeauthoryear{Alldieck, Magnor, Bhatnagar, Theobalt, and
  Pons-Moll}{Alldieck et~al\mbox{.}}{2019a}]%
        {alldieck2019learning}
\bibfield{author}{\bibinfo{person}{Thiemo Alldieck}, \bibinfo{person}{Marcus
  Magnor}, \bibinfo{person}{Bharat~Lal Bhatnagar}, \bibinfo{person}{Christian
  Theobalt}, {and} \bibinfo{person}{Gerard Pons-Moll}.}
  \bibinfo{year}{2019}\natexlab{a}.
\newblock \showarticletitle{Learning to reconstruct people in clothing from a
  single RGB camera}. In \bibinfo{booktitle}{\emph{Proceedings of the IEEE/CVF
  Conference on Computer Vision and Pattern Recognition}}.
  \bibinfo{pages}{1175--1186}.
\newblock


\bibitem[\protect\citeauthoryear{Alldieck, Magnor, Xu, Theobalt, and
  Pons-Moll}{Alldieck et~al\mbox{.}}{2018a}]%
        {alldieck2018detailed}
\bibfield{author}{\bibinfo{person}{Thiemo Alldieck}, \bibinfo{person}{Marcus
  Magnor}, \bibinfo{person}{Weipeng Xu}, \bibinfo{person}{Christian Theobalt},
  {and} \bibinfo{person}{Gerard Pons-Moll}.} \bibinfo{year}{2018}\natexlab{a}.
\newblock \showarticletitle{Detailed human avatars from monocular video}. In
  \bibinfo{booktitle}{\emph{2018 International Conference on 3D Vision (3DV)}}.
  IEEE, \bibinfo{pages}{98--109}.
\newblock


\bibitem[\protect\citeauthoryear{Alldieck, Magnor, Xu, Theobalt, and
  Pons-Moll}{Alldieck et~al\mbox{.}}{2018b}]%
        {alldieck2018video}
\bibfield{author}{\bibinfo{person}{Thiemo Alldieck}, \bibinfo{person}{Marcus
  Magnor}, \bibinfo{person}{Weipeng Xu}, \bibinfo{person}{Christian Theobalt},
  {and} \bibinfo{person}{Gerard Pons-Moll}.} \bibinfo{year}{2018}\natexlab{b}.
\newblock \showarticletitle{Video based reconstruction of 3d people models}. In
  \bibinfo{booktitle}{\emph{Proceedings of the IEEE Conference on Computer
  Vision and Pattern Recognition}}. \bibinfo{pages}{8387--8397}.
\newblock


\bibitem[\protect\citeauthoryear{Alldieck, Pons-Moll, Theobalt, and
  Magnor}{Alldieck et~al\mbox{.}}{2019b}]%
        {alldieck2019tex2shape}
\bibfield{author}{\bibinfo{person}{Thiemo Alldieck}, \bibinfo{person}{Gerard
  Pons-Moll}, \bibinfo{person}{Christian Theobalt}, {and}
  \bibinfo{person}{Marcus Magnor}.} \bibinfo{year}{2019}\natexlab{b}.
\newblock \showarticletitle{Tex2shape: Detailed full human body geometry from a
  single image}. In \bibinfo{booktitle}{\emph{Proceedings of the IEEE/CVF
  International Conference on Computer Vision}}. \bibinfo{pages}{2293--2303}.
\newblock


\bibitem[\protect\citeauthoryear{Bhatnagar, Sminchisescu, Theobalt, and
  Pons-Moll}{Bhatnagar et~al\mbox{.}}{2020}]%
        {bhatnagar2020combining}
\bibfield{author}{\bibinfo{person}{Bharat~Lal Bhatnagar},
  \bibinfo{person}{Cristian Sminchisescu}, \bibinfo{person}{Christian
  Theobalt}, {and} \bibinfo{person}{Gerard Pons-Moll}.}
  \bibinfo{year}{2020}\natexlab{}.
\newblock \showarticletitle{Combining implicit function learning and parametric
  models for 3d human reconstruction}. In \bibinfo{booktitle}{\emph{Computer
  Vision--ECCV 2020: 16th European Conference, Glasgow, UK, August 23--28,
  2020, Proceedings, Part II 16}}. Springer, \bibinfo{pages}{311--329}.
\newblock


\bibitem[\protect\citeauthoryear{Bogo, Kanazawa, Lassner, Gehler, Romero, and
  Black}{Bogo et~al\mbox{.}}{2016}]%
        {bogo2016keep}
\bibfield{author}{\bibinfo{person}{Federica Bogo}, \bibinfo{person}{Angjoo
  Kanazawa}, \bibinfo{person}{Christoph Lassner}, \bibinfo{person}{Peter
  Gehler}, \bibinfo{person}{Javier Romero}, {and} \bibinfo{person}{Michael~J
  Black}.} \bibinfo{year}{2016}\natexlab{}.
\newblock \showarticletitle{Keep it SMPL: Automatic estimation of 3D human pose
  and shape from a single image}. In \bibinfo{booktitle}{\emph{European
  conference on computer vision}}. Springer, \bibinfo{pages}{561--578}.
\newblock


\bibitem[\protect\citeauthoryear{{Cao}, {Hidalgo Martinez}, {Simon}, {Wei}, and
  {Sheikh}}{{Cao} et~al\mbox{.}}{2019}]%
        {8765346}
\bibfield{author}{\bibinfo{person}{Z. {Cao}}, \bibinfo{person}{G. {Hidalgo
  Martinez}}, \bibinfo{person}{T. {Simon}}, \bibinfo{person}{S. {Wei}}, {and}
  \bibinfo{person}{Y.~A. {Sheikh}}.} \bibinfo{year}{2019}\natexlab{}.
\newblock \showarticletitle{OpenPose: Realtime Multi-Person 2D Pose Estimation
  using Part Affinity Fields}.
\newblock \bibinfo{journal}{\emph{IEEE Transactions on Pattern Analysis and
  Machine Intelligence}} (\bibinfo{year}{2019}).
\newblock


\bibitem[\protect\citeauthoryear{Chen and Zhang}{Chen and Zhang}{2019}]%
        {chen2019learning}
\bibfield{author}{\bibinfo{person}{Zhiqin Chen} {and} \bibinfo{person}{Hao
  Zhang}.} \bibinfo{year}{2019}\natexlab{}.
\newblock \showarticletitle{Learning implicit fields for generative shape
  modeling}. In \bibinfo{booktitle}{\emph{Proceedings of the IEEE/CVF
  Conference on Computer Vision and Pattern Recognition}}.
  \bibinfo{pages}{5939--5948}.
\newblock


\bibitem[\protect\citeauthoryear{Chibane, Alldieck, and Pons-Moll}{Chibane
  et~al\mbox{.}}{2020}]%
        {chibane2020implicit}
\bibfield{author}{\bibinfo{person}{Julian Chibane}, \bibinfo{person}{Thiemo
  Alldieck}, {and} \bibinfo{person}{Gerard Pons-Moll}.}
  \bibinfo{year}{2020}\natexlab{}.
\newblock \showarticletitle{Implicit functions in feature space for 3d shape
  reconstruction and completion}. In \bibinfo{booktitle}{\emph{Proceedings of
  the IEEE/CVF Conference on Computer Vision and Pattern Recognition}}.
  \bibinfo{pages}{6970--6981}.
\newblock


\bibitem[\protect\citeauthoryear{Fan, Su, and Guibas}{Fan
  et~al\mbox{.}}{2017}]%
        {fan2017point}
\bibfield{author}{\bibinfo{person}{Haoqiang Fan}, \bibinfo{person}{Hao Su},
  {and} \bibinfo{person}{Leonidas~J Guibas}.} \bibinfo{year}{2017}\natexlab{}.
\newblock \showarticletitle{A point set generation network for 3d object
  reconstruction from a single image}. In \bibinfo{booktitle}{\emph{Proceedings
  of the IEEE conference on computer vision and pattern recognition}}.
  \bibinfo{pages}{605--613}.
\newblock


\bibitem[\protect\citeauthoryear{Gall, Stoll, De~Aguiar, Theobalt, Rosenhahn,
  and Seidel}{Gall et~al\mbox{.}}{2009}]%
        {gall2009motion}
\bibfield{author}{\bibinfo{person}{Juergen Gall}, \bibinfo{person}{Carsten
  Stoll}, \bibinfo{person}{Edilson De~Aguiar}, \bibinfo{person}{Christian
  Theobalt}, \bibinfo{person}{Bodo Rosenhahn}, {and}
  \bibinfo{person}{Hans-Peter Seidel}.} \bibinfo{year}{2009}\natexlab{}.
\newblock \showarticletitle{Motion capture using joint skeleton tracking and
  surface estimation}. In \bibinfo{booktitle}{\emph{2009 IEEE Conference on
  Computer Vision and Pattern Recognition}}. IEEE, \bibinfo{pages}{1746--1753}.
\newblock


\bibitem[\protect\citeauthoryear{Genova, Cole, Sud, Sarna, and
  Funkhouser}{Genova et~al\mbox{.}}{2020}]%
        {genova2020local}
\bibfield{author}{\bibinfo{person}{Kyle Genova}, \bibinfo{person}{Forrester
  Cole}, \bibinfo{person}{Avneesh Sud}, \bibinfo{person}{Aaron Sarna}, {and}
  \bibinfo{person}{Thomas Funkhouser}.} \bibinfo{year}{2020}\natexlab{}.
\newblock \showarticletitle{Local deep implicit functions for 3d shape}. In
  \bibinfo{booktitle}{\emph{Proceedings of the IEEE/CVF Conference on Computer
  Vision and Pattern Recognition}}. \bibinfo{pages}{4857--4866}.
\newblock


\bibitem[\protect\citeauthoryear{Girdhar, Fouhey, Rodriguez, and Gupta}{Girdhar
  et~al\mbox{.}}{2016}]%
        {girdhar2016learning}
\bibfield{author}{\bibinfo{person}{Rohit Girdhar}, \bibinfo{person}{David~F
  Fouhey}, \bibinfo{person}{Mikel Rodriguez}, {and} \bibinfo{person}{Abhinav
  Gupta}.} \bibinfo{year}{2016}\natexlab{}.
\newblock \showarticletitle{Learning a predictable and generative vector
  representation for objects}. In \bibinfo{booktitle}{\emph{European Conference
  on Computer Vision}}. Springer, \bibinfo{pages}{484--499}.
\newblock


\bibitem[\protect\citeauthoryear{Graham, Engelcke, and Van Der~Maaten}{Graham
  et~al\mbox{.}}{2018}]%
        {graham20183d}
\bibfield{author}{\bibinfo{person}{Benjamin Graham}, \bibinfo{person}{Martin
  Engelcke}, {and} \bibinfo{person}{Laurens Van Der~Maaten}.}
  \bibinfo{year}{2018}\natexlab{}.
\newblock \showarticletitle{3d semantic segmentation with submanifold sparse
  convolutional networks}. In \bibinfo{booktitle}{\emph{Proceedings of the IEEE
  conference on computer vision and pattern recognition}}.
  \bibinfo{pages}{9224--9232}.
\newblock


\bibitem[\protect\citeauthoryear{Guo, Lincoln, Davidson, Busch, Yu, Whalen,
  Harvey, Orts-Escolano, Pandey, Dourgarian, et~al\mbox{.}}{Guo
  et~al\mbox{.}}{2019}]%
        {guo2019relightables}
\bibfield{author}{\bibinfo{person}{Kaiwen Guo}, \bibinfo{person}{Peter
  Lincoln}, \bibinfo{person}{Philip Davidson}, \bibinfo{person}{Jay Busch},
  \bibinfo{person}{Xueming Yu}, \bibinfo{person}{Matt Whalen},
  \bibinfo{person}{Geoff Harvey}, \bibinfo{person}{Sergio Orts-Escolano},
  \bibinfo{person}{Rohit Pandey}, \bibinfo{person}{Jason Dourgarian},
  {et~al\mbox{.}}} \bibinfo{year}{2019}\natexlab{}.
\newblock \showarticletitle{The relightables: Volumetric performance capture of
  humans with realistic relighting}.
\newblock \bibinfo{journal}{\emph{ACM Transactions on Graphics (TOG)}}
  \bibinfo{volume}{38}, \bibinfo{number}{6} (\bibinfo{year}{2019}),
  \bibinfo{pages}{1--19}.
\newblock


\bibitem[\protect\citeauthoryear{Habermann, Xu, Zollhoefer, Pons-Moll, and
  Theobalt}{Habermann et~al\mbox{.}}{2019}]%
        {habermann2019livecap}
\bibfield{author}{\bibinfo{person}{Marc Habermann}, \bibinfo{person}{Weipeng
  Xu}, \bibinfo{person}{Michael Zollhoefer}, \bibinfo{person}{Gerard
  Pons-Moll}, {and} \bibinfo{person}{Christian Theobalt}.}
  \bibinfo{year}{2019}\natexlab{}.
\newblock \showarticletitle{Livecap: Real-time human performance capture from
  monocular video}.
\newblock \bibinfo{journal}{\emph{ACM Transactions On Graphics (TOG)}}
  \bibinfo{volume}{38}, \bibinfo{number}{2} (\bibinfo{year}{2019}),
  \bibinfo{pages}{1--17}.
\newblock


\bibitem[\protect\citeauthoryear{Habermann, Xu, Zollhofer, Pons-Moll, and
  Theobalt}{Habermann et~al\mbox{.}}{2020}]%
        {habermann2020deepcap}
\bibfield{author}{\bibinfo{person}{Marc Habermann}, \bibinfo{person}{Weipeng
  Xu}, \bibinfo{person}{Michael Zollhofer}, \bibinfo{person}{Gerard Pons-Moll},
  {and} \bibinfo{person}{Christian Theobalt}.} \bibinfo{year}{2020}\natexlab{}.
\newblock \showarticletitle{Deepcap: Monocular human performance capture using
  weak supervision}. In \bibinfo{booktitle}{\emph{Proceedings of the IEEE/CVF
  Conference on Computer Vision and Pattern Recognition}}.
  \bibinfo{pages}{5052--5063}.
\newblock


\bibitem[\protect\citeauthoryear{He, Collomosse, Jin, and Soatto}{He
  et~al\mbox{.}}{2020}]%
        {he2020geo}
\bibfield{author}{\bibinfo{person}{Tong He}, \bibinfo{person}{John Collomosse},
  \bibinfo{person}{Hailin Jin}, {and} \bibinfo{person}{Stefano Soatto}.}
  \bibinfo{year}{2020}\natexlab{}.
\newblock \showarticletitle{Geo-pifu: Geometry and pixel aligned implicit
  functions for single-view human reconstruction}.
\newblock \bibinfo{journal}{\emph{arXiv preprint arXiv:2006.08072}}
  (\bibinfo{year}{2020}).
\newblock


\bibitem[\protect\citeauthoryear{He, Xu, Saito, Soatto, and Tung}{He
  et~al\mbox{.}}{2021}]%
        {he2021arch++}
\bibfield{author}{\bibinfo{person}{Tong He}, \bibinfo{person}{Yuanlu Xu},
  \bibinfo{person}{Shunsuke Saito}, \bibinfo{person}{Stefano Soatto}, {and}
  \bibinfo{person}{Tony Tung}.} \bibinfo{year}{2021}\natexlab{}.
\newblock \showarticletitle{ARCH++: Animation-ready clothed human
  reconstruction revisited}. In \bibinfo{booktitle}{\emph{Proceedings of the
  IEEE/CVF International Conference on Computer Vision}}.
  \bibinfo{pages}{11046--11056}.
\newblock


\bibitem[\protect\citeauthoryear{Huang, Xu, Lassner, Li, and Tung}{Huang
  et~al\mbox{.}}{2020}]%
        {huang2020arch}
\bibfield{author}{\bibinfo{person}{Zeng Huang}, \bibinfo{person}{Yuanlu Xu},
  \bibinfo{person}{Christoph Lassner}, \bibinfo{person}{Hao Li}, {and}
  \bibinfo{person}{Tony Tung}.} \bibinfo{year}{2020}\natexlab{}.
\newblock \showarticletitle{Arch: Animatable reconstruction of clothed humans}.
  In \bibinfo{booktitle}{\emph{Proceedings of the IEEE/CVF Conference on
  Computer Vision and Pattern Recognition}}. \bibinfo{pages}{3093--3102}.
\newblock


\bibitem[\protect\citeauthoryear{Jiang, Sud, Makadia, Huang, Nie{\ss}ner,
  Funkhouser, et~al\mbox{.}}{Jiang et~al\mbox{.}}{2020}]%
        {jiang2020local}
\bibfield{author}{\bibinfo{person}{Chiyu Jiang}, \bibinfo{person}{Avneesh Sud},
  \bibinfo{person}{Ameesh Makadia}, \bibinfo{person}{Jingwei Huang},
  \bibinfo{person}{Matthias Nie{\ss}ner}, \bibinfo{person}{Thomas Funkhouser},
  {et~al\mbox{.}}} \bibinfo{year}{2020}\natexlab{}.
\newblock \showarticletitle{Local implicit grid representations for 3d scenes}.
  In \bibinfo{booktitle}{\emph{Proceedings of the IEEE/CVF Conference on
  Computer Vision and Pattern Recognition}}. \bibinfo{pages}{6001--6010}.
\newblock


\bibitem[\protect\citeauthoryear{Kanazawa, Black, Jacobs, and Malik}{Kanazawa
  et~al\mbox{.}}{2018}]%
        {kanazawa2018end}
\bibfield{author}{\bibinfo{person}{Angjoo Kanazawa}, \bibinfo{person}{Michael~J
  Black}, \bibinfo{person}{David~W Jacobs}, {and} \bibinfo{person}{Jitendra
  Malik}.} \bibinfo{year}{2018}\natexlab{}.
\newblock \showarticletitle{End-to-end recovery of human shape and pose}. In
  \bibinfo{booktitle}{\emph{Proceedings of the IEEE conference on computer
  vision and pattern recognition}}. \bibinfo{pages}{7122--7131}.
\newblock


\bibitem[\protect\citeauthoryear{Kocabas, Athanasiou, and Black}{Kocabas
  et~al\mbox{.}}{2020}]%
        {kocabas2020vibe}
\bibfield{author}{\bibinfo{person}{Muhammed Kocabas}, \bibinfo{person}{Nikos
  Athanasiou}, {and} \bibinfo{person}{Michael~J Black}.}
  \bibinfo{year}{2020}\natexlab{}.
\newblock \showarticletitle{Vibe: Video inference for human body pose and shape
  estimation}. In \bibinfo{booktitle}{\emph{Proceedings of the IEEE/CVF
  Conference on Computer Vision and Pattern Recognition}}.
  \bibinfo{pages}{5253--5263}.
\newblock


\bibitem[\protect\citeauthoryear{Kolotouros, Pavlakos, Black, and
  Daniilidis}{Kolotouros et~al\mbox{.}}{2019}]%
        {kolotouros2019learning}
\bibfield{author}{\bibinfo{person}{Nikos Kolotouros}, \bibinfo{person}{Georgios
  Pavlakos}, \bibinfo{person}{Michael~J Black}, {and} \bibinfo{person}{Kostas
  Daniilidis}.} \bibinfo{year}{2019}\natexlab{}.
\newblock \showarticletitle{Learning to reconstruct 3D human pose and shape via
  model-fitting in the loop}. In \bibinfo{booktitle}{\emph{Proceedings of the
  IEEE/CVF International Conference on Computer Vision}}.
  \bibinfo{pages}{2252--2261}.
\newblock


\bibitem[\protect\citeauthoryear{Lazova, Insafutdinov, and Pons-Moll}{Lazova
  et~al\mbox{.}}{2019}]%
        {lazova2019360}
\bibfield{author}{\bibinfo{person}{Verica Lazova}, \bibinfo{person}{Eldar
  Insafutdinov}, {and} \bibinfo{person}{Gerard Pons-Moll}.}
  \bibinfo{year}{2019}\natexlab{}.
\newblock \showarticletitle{360-degree textures of people in clothing from a
  single image}. In \bibinfo{booktitle}{\emph{2019 International Conference on
  3D Vision (3DV)}}. IEEE, \bibinfo{pages}{643--653}.
\newblock


\bibitem[\protect\citeauthoryear{Leroy, Franco, and Boyer}{Leroy
  et~al\mbox{.}}{2017}]%
        {leroy2017multi}
\bibfield{author}{\bibinfo{person}{Vincent Leroy},
  \bibinfo{person}{Jean-S{\'e}bastien Franco}, {and} \bibinfo{person}{Edmond
  Boyer}.} \bibinfo{year}{2017}\natexlab{}.
\newblock \showarticletitle{Multi-view dynamic shape refinement using local
  temporal integration}. In \bibinfo{booktitle}{\emph{Proceedings of the IEEE
  international conference on computer vision}}. \bibinfo{pages}{3094--3103}.
\newblock


\bibitem[\protect\citeauthoryear{Li, Yu, Zheng, Guo, and Liu}{Li
  et~al\mbox{.}}{2021}]%
        {li2021posefusion}
\bibfield{author}{\bibinfo{person}{Zhe Li}, \bibinfo{person}{Tao Yu},
  \bibinfo{person}{Zerong Zheng}, \bibinfo{person}{Kaiwen Guo}, {and}
  \bibinfo{person}{Yebin Liu}.} \bibinfo{year}{2021}\natexlab{}.
\newblock \showarticletitle{POSEFusion: Pose-guided Selective Fusion for
  Single-view Human Volumetric Capture}. In
  \bibinfo{booktitle}{\emph{Proceedings of the IEEE/CVF Conference on Computer
  Vision and Pattern Recognition}}. \bibinfo{pages}{14162--14172}.
\newblock


\bibitem[\protect\citeauthoryear{Lin, Kong, and Lucey}{Lin
  et~al\mbox{.}}{2018}]%
        {lin2018learning}
\bibfield{author}{\bibinfo{person}{Chen-Hsuan Lin}, \bibinfo{person}{Chen
  Kong}, {and} \bibinfo{person}{Simon Lucey}.} \bibinfo{year}{2018}\natexlab{}.
\newblock \showarticletitle{Learning efficient point cloud generation for dense
  3d object reconstruction}. In \bibinfo{booktitle}{\emph{proceedings of the
  AAAI Conference on Artificial Intelligence}}, Vol.~\bibinfo{volume}{32}.
\newblock


\bibitem[\protect\citeauthoryear{Lin, Ryabtsev, Sengupta, Curless, Seitz, and
  Kemelmacher-Shlizerman}{Lin et~al\mbox{.}}{2020}]%
        {BGMv2}
\bibfield{author}{\bibinfo{person}{Shanchuan Lin}, \bibinfo{person}{Andrey
  Ryabtsev}, \bibinfo{person}{Soumyadip Sengupta}, \bibinfo{person}{Brian
  Curless}, \bibinfo{person}{Steve Seitz}, {and} \bibinfo{person}{Ira
  Kemelmacher-Shlizerman}.} \bibinfo{year}{2020}\natexlab{}.
\newblock \showarticletitle{Real-Time High-Resolution Background Matting}.
\newblock \bibinfo{journal}{\emph{arXiv}} (\bibinfo{year}{2020}),
  \bibinfo{pages}{arXiv--2012}.
\newblock


\bibitem[\protect\citeauthoryear{Liu, Gall, Stoll, Dai, Seidel, and
  Theobalt}{Liu et~al\mbox{.}}{2013}]%
        {liu2013markerless}
\bibfield{author}{\bibinfo{person}{Yebin Liu}, \bibinfo{person}{Juergen Gall},
  \bibinfo{person}{Carsten Stoll}, \bibinfo{person}{Qionghai Dai},
  \bibinfo{person}{Hans-Peter Seidel}, {and} \bibinfo{person}{Christian
  Theobalt}.} \bibinfo{year}{2013}\natexlab{}.
\newblock \showarticletitle{Markerless motion capture of multiple characters
  using multiview image segmentation}.
\newblock \bibinfo{journal}{\emph{IEEE transactions on pattern analysis and
  machine intelligence}} \bibinfo{volume}{35}, \bibinfo{number}{11}
  (\bibinfo{year}{2013}), \bibinfo{pages}{2720--2735}.
\newblock


\bibitem[\protect\citeauthoryear{Lombardi, Simon, Saragih, Schwartz, Lehrmann,
  and Sheikh}{Lombardi et~al\mbox{.}}{2019}]%
        {lombardi2019neural}
\bibfield{author}{\bibinfo{person}{Stephen Lombardi}, \bibinfo{person}{Tomas
  Simon}, \bibinfo{person}{Jason Saragih}, \bibinfo{person}{Gabriel Schwartz},
  \bibinfo{person}{Andreas Lehrmann}, {and} \bibinfo{person}{Yaser Sheikh}.}
  \bibinfo{year}{2019}\natexlab{}.
\newblock \showarticletitle{Neural volumes: Learning dynamic renderable volumes
  from images}.
\newblock \bibinfo{journal}{\emph{arXiv preprint arXiv:1906.07751}}
  (\bibinfo{year}{2019}).
\newblock


\bibitem[\protect\citeauthoryear{Loper, Mahmood, Romero, Pons-Moll, and
  Black}{Loper et~al\mbox{.}}{2015}]%
        {loper2015smpl}
\bibfield{author}{\bibinfo{person}{Matthew Loper}, \bibinfo{person}{Naureen
  Mahmood}, \bibinfo{person}{Javier Romero}, \bibinfo{person}{Gerard
  Pons-Moll}, {and} \bibinfo{person}{Michael~J Black}.}
  \bibinfo{year}{2015}\natexlab{}.
\newblock \showarticletitle{SMPL: A skinned multi-person linear model}.
\newblock \bibinfo{journal}{\emph{ACM transactions on graphics (TOG)}}
  \bibinfo{volume}{34}, \bibinfo{number}{6} (\bibinfo{year}{2015}),
  \bibinfo{pages}{1--16}.
\newblock


\bibitem[\protect\citeauthoryear{Lorensen and Cline}{Lorensen and
  Cline}{1987}]%
        {lorensen1987marching}
\bibfield{author}{\bibinfo{person}{William~E Lorensen} {and}
  \bibinfo{person}{Harvey~E Cline}.} \bibinfo{year}{1987}\natexlab{}.
\newblock \showarticletitle{Marching cubes: A high resolution 3D surface
  construction algorithm}.
\newblock \bibinfo{journal}{\emph{ACM siggraph computer graphics}}
  \bibinfo{volume}{21}, \bibinfo{number}{4} (\bibinfo{year}{1987}),
  \bibinfo{pages}{163--169}.
\newblock


\bibitem[\protect\citeauthoryear{Ma, Yang, Ranjan, Pujades, Pons-Moll, Tang,
  and Black}{Ma et~al\mbox{.}}{2020}]%
        {ma2020learning}
\bibfield{author}{\bibinfo{person}{Qianli Ma}, \bibinfo{person}{Jinlong Yang},
  \bibinfo{person}{Anurag Ranjan}, \bibinfo{person}{Sergi Pujades},
  \bibinfo{person}{Gerard Pons-Moll}, \bibinfo{person}{Siyu Tang}, {and}
  \bibinfo{person}{Michael~J Black}.} \bibinfo{year}{2020}\natexlab{}.
\newblock \showarticletitle{Learning to dress 3d people in generative
  clothing}. In \bibinfo{booktitle}{\emph{Proceedings of the IEEE/CVF
  Conference on Computer Vision and Pattern Recognition}}.
  \bibinfo{pages}{6469--6478}.
\newblock


\bibitem[\protect\citeauthoryear{Mustafa, Kim, Guillemaut, and Hilton}{Mustafa
  et~al\mbox{.}}{2015}]%
        {mustafa2015general}
\bibfield{author}{\bibinfo{person}{Armin Mustafa}, \bibinfo{person}{Hansung
  Kim}, \bibinfo{person}{Jean-Yves Guillemaut}, {and} \bibinfo{person}{Adrian
  Hilton}.} \bibinfo{year}{2015}\natexlab{}.
\newblock \showarticletitle{General dynamic scene reconstruction from multiple
  view video}. In \bibinfo{booktitle}{\emph{Proceedings of the IEEE
  International Conference on Computer Vision}}. \bibinfo{pages}{900--908}.
\newblock


\bibitem[\protect\citeauthoryear{Newcombe, Fox, and Seitz}{Newcombe
  et~al\mbox{.}}{2015}]%
        {newcombe2015dynamicfusion}
\bibfield{author}{\bibinfo{person}{Richard~A Newcombe}, \bibinfo{person}{Dieter
  Fox}, {and} \bibinfo{person}{Steven~M Seitz}.}
  \bibinfo{year}{2015}\natexlab{}.
\newblock \showarticletitle{Dynamicfusion: Reconstruction and tracking of
  non-rigid scenes in real-time}. In \bibinfo{booktitle}{\emph{Proceedings of
  the IEEE conference on computer vision and pattern recognition}}.
  \bibinfo{pages}{343--352}.
\newblock


\bibitem[\protect\citeauthoryear{Osman, Bolkart, and Black}{Osman
  et~al\mbox{.}}{2020}]%
        {osman2020star}
\bibfield{author}{\bibinfo{person}{Ahmed~AA Osman}, \bibinfo{person}{Timo
  Bolkart}, {and} \bibinfo{person}{Michael~J Black}.}
  \bibinfo{year}{2020}\natexlab{}.
\newblock \showarticletitle{Star: Sparse trained articulated human body
  regressor}. In \bibinfo{booktitle}{\emph{Computer Vision--ECCV 2020: 16th
  European Conference, Glasgow, UK, August 23--28, 2020, Proceedings, Part VI
  16}}. Springer, \bibinfo{pages}{598--613}.
\newblock


\bibitem[\protect\citeauthoryear{Park, Florence, Straub, Newcombe, and
  Lovegrove}{Park et~al\mbox{.}}{2019}]%
        {park2019deepsdf}
\bibfield{author}{\bibinfo{person}{Jeong~Joon Park}, \bibinfo{person}{Peter
  Florence}, \bibinfo{person}{Julian Straub}, \bibinfo{person}{Richard
  Newcombe}, {and} \bibinfo{person}{Steven Lovegrove}.}
  \bibinfo{year}{2019}\natexlab{}.
\newblock \showarticletitle{Deepsdf: Learning continuous signed distance
  functions for shape representation}. In \bibinfo{booktitle}{\emph{Proceedings
  of the IEEE/CVF Conference on Computer Vision and Pattern Recognition}}.
  \bibinfo{pages}{165--174}.
\newblock


\bibitem[\protect\citeauthoryear{Peng, Niemeyer, Mescheder, Pollefeys, and
  Geiger}{Peng et~al\mbox{.}}{2020}]%
        {peng2020convolutional}
\bibfield{author}{\bibinfo{person}{Songyou Peng}, \bibinfo{person}{Michael
  Niemeyer}, \bibinfo{person}{Lars Mescheder}, \bibinfo{person}{Marc
  Pollefeys}, {and} \bibinfo{person}{Andreas Geiger}.}
  \bibinfo{year}{2020}\natexlab{}.
\newblock \showarticletitle{Convolutional occupancy networks}. In
  \bibinfo{booktitle}{\emph{Computer Vision--ECCV 2020: 16th European
  Conference, Glasgow, UK, August 23--28, 2020, Proceedings, Part III 16}}.
  Springer, \bibinfo{pages}{523--540}.
\newblock


\bibitem[\protect\citeauthoryear{Peng, Zhang, Xu, Wang, Shuai, Bao, and
  Zhou}{Peng et~al\mbox{.}}{2021}]%
        {peng2021neural}
\bibfield{author}{\bibinfo{person}{Sida Peng}, \bibinfo{person}{Yuanqing
  Zhang}, \bibinfo{person}{Yinghao Xu}, \bibinfo{person}{Qianqian Wang},
  \bibinfo{person}{Qing Shuai}, \bibinfo{person}{Hujun Bao}, {and}
  \bibinfo{person}{Xiaowei Zhou}.} \bibinfo{year}{2021}\natexlab{}.
\newblock \showarticletitle{Neural body: Implicit neural representations with
  structured latent codes for novel view synthesis of dynamic humans}. In
  \bibinfo{booktitle}{\emph{Proceedings of the IEEE/CVF Conference on Computer
  Vision and Pattern Recognition}}. \bibinfo{pages}{9054--9063}.
\newblock


\bibitem[\protect\citeauthoryear{Qi, Su, Mo, and Guibas}{Qi
  et~al\mbox{.}}{2017}]%
        {qi2017pointnet}
\bibfield{author}{\bibinfo{person}{Charles~R Qi}, \bibinfo{person}{Hao Su},
  \bibinfo{person}{Kaichun Mo}, {and} \bibinfo{person}{Leonidas~J Guibas}.}
  \bibinfo{year}{2017}\natexlab{}.
\newblock \showarticletitle{Pointnet: Deep learning on point sets for 3d
  classification and segmentation}. In \bibinfo{booktitle}{\emph{Proceedings of
  the IEEE conference on computer vision and pattern recognition}}.
  \bibinfo{pages}{652--660}.
\newblock


\bibitem[\protect\citeauthoryear{renderpeople}{renderpeople}{2000}]%
        {renderpeople}
\bibfield{author}{\bibinfo{person}{renderpeople}.}
  \bibinfo{year}{2000}\natexlab{}.
\newblock \bibinfo{howpublished}{\url{https://renderpeople.com/}}.
\newblock


\bibitem[\protect\citeauthoryear{Saito, Huang, Natsume, Morishima, Kanazawa,
  and Li}{Saito et~al\mbox{.}}{2019}]%
        {saito2019pifu}
\bibfield{author}{\bibinfo{person}{Shunsuke Saito}, \bibinfo{person}{Zeng
  Huang}, \bibinfo{person}{Ryota Natsume}, \bibinfo{person}{Shigeo Morishima},
  \bibinfo{person}{Angjoo Kanazawa}, {and} \bibinfo{person}{Hao Li}.}
  \bibinfo{year}{2019}\natexlab{}.
\newblock \showarticletitle{Pifu: Pixel-aligned implicit function for
  high-resolution clothed human digitization}. In
  \bibinfo{booktitle}{\emph{Proceedings of the IEEE/CVF International
  Conference on Computer Vision}}. \bibinfo{pages}{2304--2314}.
\newblock


\bibitem[\protect\citeauthoryear{Saito, Simon, Saragih, and Joo}{Saito
  et~al\mbox{.}}{2020}]%
        {saito2020pifuhd}
\bibfield{author}{\bibinfo{person}{Shunsuke Saito}, \bibinfo{person}{Tomas
  Simon}, \bibinfo{person}{Jason Saragih}, {and} \bibinfo{person}{Hanbyul
  Joo}.} \bibinfo{year}{2020}\natexlab{}.
\newblock \showarticletitle{Pifuhd: Multi-level pixel-aligned implicit function
  for high-resolution 3d human digitization}. In
  \bibinfo{booktitle}{\emph{Proceedings of the IEEE/CVF Conference on Computer
  Vision and Pattern Recognition}}. \bibinfo{pages}{84--93}.
\newblock


\bibitem[\protect\citeauthoryear{Shao, Zhang, Zhang, Cao, Yu, and Liu}{Shao
  et~al\mbox{.}}{2021}]%
        {shao2021doublefield}
\bibfield{author}{\bibinfo{person}{Ruizhi Shao}, \bibinfo{person}{Hongwen
  Zhang}, \bibinfo{person}{He Zhang}, \bibinfo{person}{Yanpei Cao},
  \bibinfo{person}{Tao Yu}, {and} \bibinfo{person}{Yebin Liu}.}
  \bibinfo{year}{2021}\natexlab{}.
\newblock \showarticletitle{DoubleField: Bridging the Neural Surface and
  Radiance Fields for High-fidelity Human Rendering}.
\newblock \bibinfo{journal}{\emph{arXiv preprint arXiv:2106.03798}}
  (\bibinfo{year}{2021}).
\newblock


\bibitem[\protect\citeauthoryear{Shi, Guo, Jiang, Wang, Shi, Wang, and Li}{Shi
  et~al\mbox{.}}{2020}]%
        {shi2020pv}
\bibfield{author}{\bibinfo{person}{Shaoshuai Shi}, \bibinfo{person}{Chaoxu
  Guo}, \bibinfo{person}{Li Jiang}, \bibinfo{person}{Zhe Wang},
  \bibinfo{person}{Jianping Shi}, \bibinfo{person}{Xiaogang Wang}, {and}
  \bibinfo{person}{Hongsheng Li}.} \bibinfo{year}{2020}\natexlab{}.
\newblock \showarticletitle{Pv-rcnn: Point-voxel feature set abstraction for 3d
  object detection}. In \bibinfo{booktitle}{\emph{Proceedings of the IEEE/CVF
  Conference on Computer Vision and Pattern Recognition}}.
  \bibinfo{pages}{10529--10538}.
\newblock


\bibitem[\protect\citeauthoryear{Sitzmann, Thies, Heide, Nie{\ss}ner,
  Wetzstein, and Zollhofer}{Sitzmann et~al\mbox{.}}{2019}]%
        {sitzmann2019deepvoxels}
\bibfield{author}{\bibinfo{person}{Vincent Sitzmann}, \bibinfo{person}{Justus
  Thies}, \bibinfo{person}{Felix Heide}, \bibinfo{person}{Matthias
  Nie{\ss}ner}, \bibinfo{person}{Gordon Wetzstein}, {and}
  \bibinfo{person}{Michael Zollhofer}.} \bibinfo{year}{2019}\natexlab{}.
\newblock \showarticletitle{Deepvoxels: Learning persistent 3d feature
  embeddings}. In \bibinfo{booktitle}{\emph{Proceedings of the IEEE/CVF
  Conference on Computer Vision and Pattern Recognition}}.
  \bibinfo{pages}{2437--2446}.
\newblock


\bibitem[\protect\citeauthoryear{Sun, Xiao, Liu, and Wang}{Sun
  et~al\mbox{.}}{2019}]%
        {sun2019deep}
\bibfield{author}{\bibinfo{person}{Ke Sun}, \bibinfo{person}{Bin Xiao},
  \bibinfo{person}{Dong Liu}, {and} \bibinfo{person}{Jingdong Wang}.}
  \bibinfo{year}{2019}\natexlab{}.
\newblock \showarticletitle{Deep high-resolution representation learning for
  human pose estimation}. In \bibinfo{booktitle}{\emph{Proceedings of the
  IEEE/CVF Conference on Computer Vision and Pattern Recognition}}.
  \bibinfo{pages}{5693--5703}.
\newblock


\bibitem[\protect\citeauthoryear{Xu, Cheng, Guo, Han, Liu, and Fang}{Xu
  et~al\mbox{.}}{2019}]%
        {xu2019flyfusion}
\bibfield{author}{\bibinfo{person}{Lan Xu}, \bibinfo{person}{Wei Cheng},
  \bibinfo{person}{Kaiwen Guo}, \bibinfo{person}{Lei Han},
  \bibinfo{person}{Yebin Liu}, {and} \bibinfo{person}{Lu Fang}.}
  \bibinfo{year}{2019}\natexlab{}.
\newblock \showarticletitle{Flyfusion: Realtime dynamic scene reconstruction
  using a flying depth camera}.
\newblock \bibinfo{journal}{\emph{IEEE transactions on visualization and
  computer graphics}} \bibinfo{volume}{27}, \bibinfo{number}{1}
  (\bibinfo{year}{2019}), \bibinfo{pages}{68--82}.
\newblock


\bibitem[\protect\citeauthoryear{Yan, Yang, Yumer, Guo, and Lee}{Yan
  et~al\mbox{.}}{2016}]%
        {yan2016perspective}
\bibfield{author}{\bibinfo{person}{Xinchen Yan}, \bibinfo{person}{Jimei Yang},
  \bibinfo{person}{Ersin Yumer}, \bibinfo{person}{Yijie Guo}, {and}
  \bibinfo{person}{Honglak Lee}.} \bibinfo{year}{2016}\natexlab{}.
\newblock \showarticletitle{Perspective transformer nets: Learning single-view
  3d object reconstruction without 3d supervision}.
\newblock \bibinfo{journal}{\emph{arXiv preprint arXiv:1612.00814}}
  (\bibinfo{year}{2016}).
\newblock


\bibitem[\protect\citeauthoryear{Yu, Ye, Tancik, and Kanazawa}{Yu
  et~al\mbox{.}}{2021a}]%
        {yu2021pixelnerf}
\bibfield{author}{\bibinfo{person}{Alex Yu}, \bibinfo{person}{Vickie Ye},
  \bibinfo{person}{Matthew Tancik}, {and} \bibinfo{person}{Angjoo Kanazawa}.}
  \bibinfo{year}{2021}\natexlab{a}.
\newblock \showarticletitle{pixelnerf: Neural radiance fields from one or few
  images}. In \bibinfo{booktitle}{\emph{Proceedings of the IEEE/CVF Conference
  on Computer Vision and Pattern Recognition}}. \bibinfo{pages}{4578--4587}.
\newblock


\bibitem[\protect\citeauthoryear{Yu, Guo, Xu, Dong, Su, Zhao, Li, Dai, and
  Liu}{Yu et~al\mbox{.}}{2017}]%
        {yu2017bodyfusion}
\bibfield{author}{\bibinfo{person}{Tao Yu}, \bibinfo{person}{Kaiwen Guo},
  \bibinfo{person}{Feng Xu}, \bibinfo{person}{Yuan Dong},
  \bibinfo{person}{Zhaoqi Su}, \bibinfo{person}{Jianhui Zhao},
  \bibinfo{person}{Jianguo Li}, \bibinfo{person}{Qionghai Dai}, {and}
  \bibinfo{person}{Yebin Liu}.} \bibinfo{year}{2017}\natexlab{}.
\newblock \showarticletitle{Bodyfusion: Real-time capture of human motion and
  surface geometry using a single depth camera}. In
  \bibinfo{booktitle}{\emph{Proceedings of the IEEE International Conference on
  Computer Vision}}. \bibinfo{pages}{910--919}.
\newblock


\bibitem[\protect\citeauthoryear{Yu, Zheng, Guo, Liu, Dai, and Liu}{Yu
  et~al\mbox{.}}{2021b}]%
        {yu2021function4d}
\bibfield{author}{\bibinfo{person}{Tao Yu}, \bibinfo{person}{Zerong Zheng},
  \bibinfo{person}{Kaiwen Guo}, \bibinfo{person}{Pengpeng Liu},
  \bibinfo{person}{Qionghai Dai}, {and} \bibinfo{person}{Yebin Liu}.}
  \bibinfo{year}{2021}\natexlab{b}.
\newblock \showarticletitle{Function4D: Real-time Human Volumetric Capture from
  Very Sparse Consumer RGBD Sensors}. In \bibinfo{booktitle}{\emph{Proceedings
  of the IEEE/CVF Conference on Computer Vision and Pattern Recognition}}.
  \bibinfo{pages}{5746--5756}.
\newblock


\bibitem[\protect\citeauthoryear{Yu, Zheng, Guo, Zhao, Dai, Li, Pons-Moll, and
  Liu}{Yu et~al\mbox{.}}{2018}]%
        {yu2018doublefusion}
\bibfield{author}{\bibinfo{person}{Tao Yu}, \bibinfo{person}{Zerong Zheng},
  \bibinfo{person}{Kaiwen Guo}, \bibinfo{person}{Jianhui Zhao},
  \bibinfo{person}{Qionghai Dai}, \bibinfo{person}{Hao Li},
  \bibinfo{person}{Gerard Pons-Moll}, {and} \bibinfo{person}{Yebin Liu}.}
  \bibinfo{year}{2018}\natexlab{}.
\newblock \showarticletitle{Doublefusion: Real-time capture of human
  performances with inner body shapes from a single depth sensor}. In
  \bibinfo{booktitle}{\emph{Proceedings of the IEEE conference on computer
  vision and pattern recognition}}. \bibinfo{pages}{7287--7296}.
\newblock


\bibitem[\protect\citeauthoryear{Zeng, Ouyang, Luo, Liu, and Wang}{Zeng
  et~al\mbox{.}}{2020}]%
        {zeng20203d}
\bibfield{author}{\bibinfo{person}{Wang Zeng}, \bibinfo{person}{Wanli Ouyang},
  \bibinfo{person}{Ping Luo}, \bibinfo{person}{Wentao Liu}, {and}
  \bibinfo{person}{Xiaogang Wang}.} \bibinfo{year}{2020}\natexlab{}.
\newblock \showarticletitle{3d human mesh regression with dense
  correspondence}. In \bibinfo{booktitle}{\emph{Proceedings of the IEEE/CVF
  Conference on Computer Vision and Pattern Recognition}}.
  \bibinfo{pages}{7054--7063}.
\newblock


\bibitem[\protect\citeauthoryear{Zhang, Pujades, Black, and Pons-Moll}{Zhang
  et~al\mbox{.}}{2017}]%
        {zhang2017detailed}
\bibfield{author}{\bibinfo{person}{Chao Zhang}, \bibinfo{person}{Sergi
  Pujades}, \bibinfo{person}{Michael~J Black}, {and} \bibinfo{person}{Gerard
  Pons-Moll}.} \bibinfo{year}{2017}\natexlab{}.
\newblock \showarticletitle{Detailed, accurate, human shape estimation from
  clothed 3D scan sequences}. In \bibinfo{booktitle}{\emph{Proceedings of the
  IEEE Conference on Computer Vision and Pattern Recognition}}.
  \bibinfo{pages}{4191--4200}.
\newblock


\bibitem[\protect\citeauthoryear{Zhang, Tian, Zhou, Ouyang, Liu, Wang, and
  Sun}{Zhang et~al\mbox{.}}{2021}]%
        {zhang2021pymaf}
\bibfield{author}{\bibinfo{person}{Hongwen Zhang}, \bibinfo{person}{Yating
  Tian}, \bibinfo{person}{Xinchi Zhou}, \bibinfo{person}{Wanli Ouyang},
  \bibinfo{person}{Yebin Liu}, \bibinfo{person}{Limin Wang}, {and}
  \bibinfo{person}{Zhenan Sun}.} \bibinfo{year}{2021}\natexlab{}.
\newblock \showarticletitle{Pymaf: 3d human pose and shape regression with
  pyramidal mesh alignment feedback loop}. In
  \bibinfo{booktitle}{\emph{Proceedings of the IEEE/CVF International
  Conference on Computer Vision}}. \bibinfo{pages}{11446--11456}.
\newblock


\bibitem[\protect\citeauthoryear{Zheng, Shao, Zhang, Yu, Zheng, Dai, and
  Liu}{Zheng et~al\mbox{.}}{2021a}]%
        {zheng2021deepmulticap}
\bibfield{author}{\bibinfo{person}{Yang Zheng}, \bibinfo{person}{Ruizhi Shao},
  \bibinfo{person}{Yuxiang Zhang}, \bibinfo{person}{Tao Yu},
  \bibinfo{person}{Zerong Zheng}, \bibinfo{person}{Qionghai Dai}, {and}
  \bibinfo{person}{Yebin Liu}.} \bibinfo{year}{2021}\natexlab{a}.
\newblock \showarticletitle{DeepMultiCap: Performance Capture of Multiple
  Characters Using Sparse Multiview Cameras}.
\newblock \bibinfo{journal}{\emph{arXiv preprint arXiv:2105.00261}}
  (\bibinfo{year}{2021}).
\newblock


\bibitem[\protect\citeauthoryear{Zheng, Yu, Liu, and Dai}{Zheng
  et~al\mbox{.}}{2021b}]%
        {zheng2021pamir}
\bibfield{author}{\bibinfo{person}{Zerong Zheng}, \bibinfo{person}{Tao Yu},
  \bibinfo{person}{Yebin Liu}, {and} \bibinfo{person}{Qionghai Dai}.}
  \bibinfo{year}{2021}\natexlab{b}.
\newblock \showarticletitle{Pamir: Parametric model-conditioned implicit
  representation for image-based human reconstruction}.
\newblock \bibinfo{journal}{\emph{IEEE Transactions on Pattern Analysis and
  Machine Intelligence}} (\bibinfo{year}{2021}).
\newblock


\bibitem[\protect\citeauthoryear{Zhi, Lassner, Tung, Stoll, Narasimhan, and
  Vo}{Zhi et~al\mbox{.}}{2020}]%
        {zhi2020texmesh}
\bibfield{author}{\bibinfo{person}{Tiancheng Zhi}, \bibinfo{person}{Christoph
  Lassner}, \bibinfo{person}{Tony Tung}, \bibinfo{person}{Carsten Stoll},
  \bibinfo{person}{Srinivasa~G Narasimhan}, {and} \bibinfo{person}{Minh Vo}.}
  \bibinfo{year}{2020}\natexlab{}.
\newblock \showarticletitle{Texmesh: Reconstructing detailed human texture and
  geometry from rgb-d video}. In \bibinfo{booktitle}{\emph{European Conference
  on Computer Vision}}. Springer, \bibinfo{pages}{492--509}.
\newblock


\end{thebibliography}
\clearpage
%%
%% If your work has an appendix, this is the place to put it.
\newpage
\appendix

\section{Appendix}

\subsection{Generation of Ground Truth SMPL}
% no \IEEEPARstart
In our experiments, we register SMPL models to 3D scans in our dataset to obtain accurate SPML parameters and corresponding 3D models. The registered SMPL models are described as ground truth SMPL in our paper. To achieve reliable registrations, we optimize SMPL parameters iteratively by minimizing the chamfer loss between SMPL vertices and scan vertices together with 2D joints loss between semantic-aligned SMPL body joints projected on images and 2D keypoints extracted by Openpose~\cite{8765346}.

% \newpage

\begin{figure*}[b] %H为当前位置，!htb为忽略美学标准，htbp为浮动图形
\centering %图片居中
\includegraphics[width=1\textwidth,trim=0 120 70 0,clip]{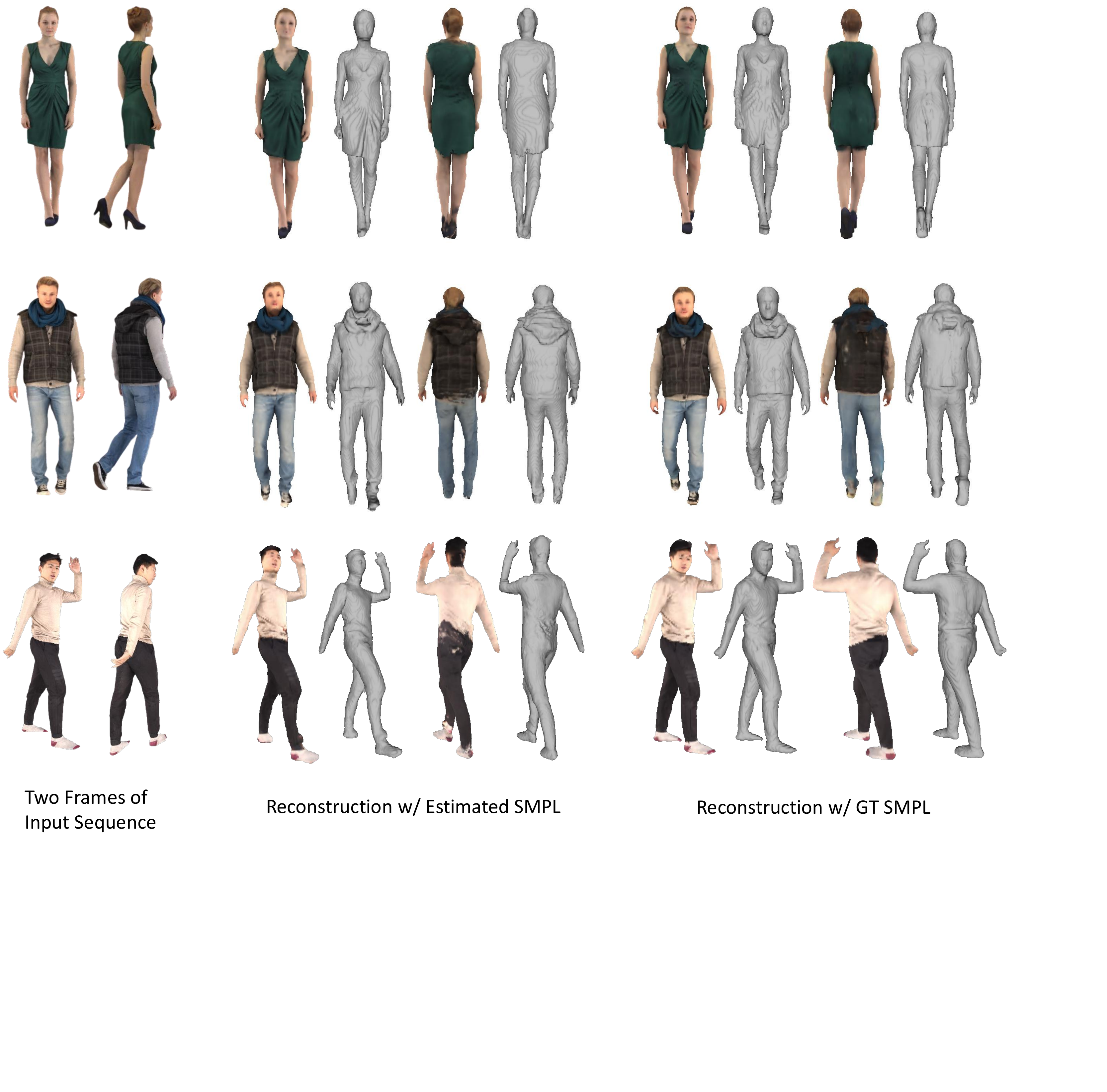} %插入图片，[]中设置图片大小，{}中是图片文件名
\caption{More results on synthetic data} %最终文档中希望显示的图片标题
\label{supp} %用于文内引用的标签
\end{figure*}

\subsection{Additional Results}
\noindent \textbf{Additional Qualitative Results} We present additional qualitative results in Fig.\ref{supp} to further demonstrate the effectiveness of our CrossHuman.

\begin{figure*}[b] %H为当前位置，!htb为忽略美学标准，htbp为浮动图形
\centering %图片居中
\includegraphics[width=1\textwidth,trim=0 0 60 0,clip]{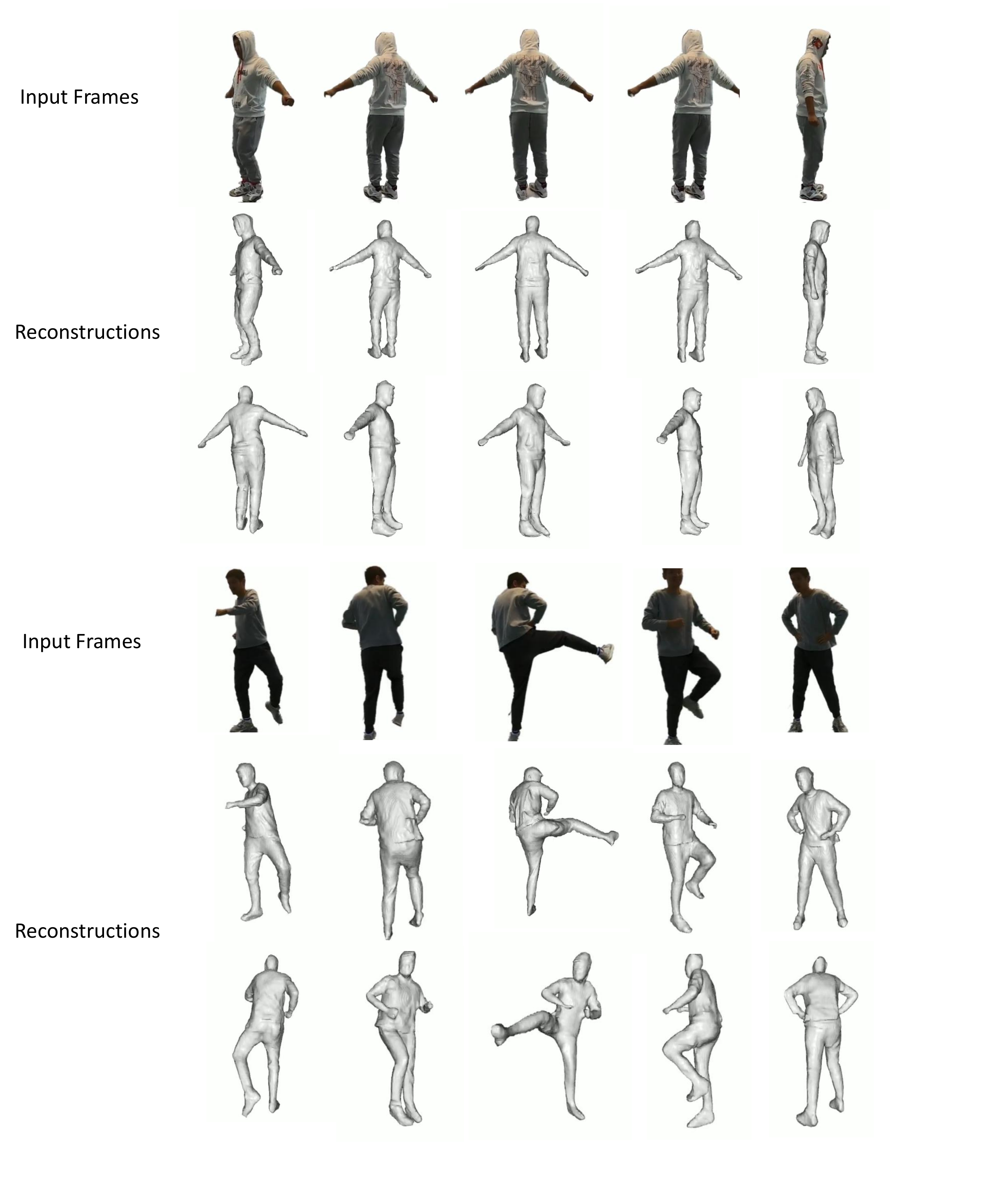} %插入图片，[]中设置图片大小，{}中是图片文件名
\caption{Results on Real Video} %最终文档中希望显示的图片标题
\label{suppv} %用于文内引用的标签
\end{figure*}

\noindent \textbf{Video Results} We provide supplemental videos of dynamic reconstruction from self-captured videos. Results on some of the frames are presented in Fig.~\ref{suppv}. The backgrounds of videos are removed by BackgroundMattingV2~\cite{BGMv2}.

\end{document}